\documentclass[twoside,11pt,abbrvbib]{article}
\usepackage{Preamble}

\usepackage{fancyhdr}

\fancypagestyle{firstpage} {%
   \fancyfoot[C]{\emph{Journal of Machine Learning Research, accepted subject to minor revisions.}}}

\title{{\bf Evolutionary Variational Optimization of\\Generative Models}}

\author{
\large
Jakob Drefs$^{1}$, Enrico Guiraud$^{1,2}$, J\"org L\"ucke$^{1}$ \\
\normalsize jakob.drefs@uol.de, enrico.guiraud@cern.ch, joerg.luecke@uol.de \\[2mm]
\normalsize $^1$\,Machine Learning Lab, University of Oldenburg, Germany \\
\normalsize $^2$\,CERN, Switzerland
\vspace{20pt}
}

\date{}

\begin{document}

\maketitle

\thispagestyle{firstpage}
\enlargethispage{-5\baselineskip}  

\begin{abstract}
\noindent
We combine two popular optimization approaches to derive learning algorithms for generative models: variational optimization and evolutionary algorithms.
The combination is realized for generative models with discrete latents by using truncated posteriors as the family of variational distributions. The variational parameters
of truncated posteriors are sets of latent states. By interpreting these states as genomes of individuals and by using the variational lower bound to define a fitness, we can
apply evolutionary algorithms to realize the variational loop. The used variational distributions are very flexible and we show that evolutionary algorithms can effectively and
efficiently optimize the variational bound. Furthermore, the variational loop is generally applicable (``black box'') with no analytical derivations required.
To show general applicability, we apply the approach to three generative models (we use noisy-OR Bayes Nets, Binary Sparse Coding, and Spike-and-Slab Sparse Coding).
To demonstrate effectiveness and efficiency of the novel variational approach, we use the standard competitive benchmarks of image denoising and inpainting. 
The benchmarks allow quantitative comparisons to a wide range of methods including probabilistic approaches, deep deterministic and generative networks, and non-local
image processing methods.
%
%
In the category of ``zero-shot'' learning (when only the corrupted image is used for training), we observed the evolutionary variational algorithm to significantly improve
the state-of-the-art in many benchmark settings.
%
%
For one well-known inpainting benchmark, we also observed state-of-the-art performance across all categories of algorithms although we only train on
the corrupted image.
In general, our investigations highlight the importance of research on optimization methods for generative models to achieve performance improvements.
%
%
%
%
%
%
%
%
%
\end{abstract}


\section{Introduction}
Variational approximations \citep[][and many more]{SaulEtAl1996,SaulJordan1996,NealHinton1998,JordanEtAl1999} are very popular and successful approaches to efficiently train probabilistic generative models. In the very common case when inference based on a generative data model is not tractable or not scalable to desired problem sizes, variational approaches provide approximations for an efficient optimization of model parameters. 
Here we focus on variational expectation maximization (variational EM) to derive learning algorithms \citep[][]{NealHinton1998,JordanEtAl1999}, while acknowledging that variational approaches are also well suited and very successful in the context of fully Bayesian (non-parametric) approaches \citep[e.g.][]{GhahramaniJordan1995,ZhouEtAl2009}. Variational EM typically seeks to approximate intractable full posterior distributions by members of a specific family of variational distributions. Prominent such families are the family of Gaussian distributions \citep[e.g.,][]{OpperWinther2005,RezendeEtAl2014,KingmaWelling2014} and the family of factored distributions \citep[e.g.,][]{JordanEtAl1999} with the latter often being referred to as {\em mean field} approaches. Variational approaches can, in general, be regarded as one of two large classes of approximations with the other being made up of sampling approximations. Variational training can be very efficient and has been shown to realize training of very large-scale models \citep[][]{BleiEtAl2003,RezendeEtAl2014,KingmaWelling2014} with thousands or even millions \citep[][]{SheikhLucke2016} of parameters.
In terms of efficiency and scalability they are often preferred to sampling approaches \citep[see, e.g.,][Section\,6, for a discussion]{AngelinoEtAl2016}, although many successful algorithms often combine both variational and sampling techniques \citep[e.g.][]{DayanEtAl1995,SheltonEtAl2011,RezendeEtAl2014,KingmaWelling2014}. 

One drawback of variational approaches lies in their limits in modeling the posterior structure, which can result in biases introduced into the learned parameters. Gaussian variational distributions, by definition, only capture one posterior mode, for instance, and fully factored approaches (i.e., mean field) do not capture explaining-away effects including posterior correlations. Gaussian variational EM is consequently particularly popular for generative models known to result in posteriors with essentially one mode \citep[][]{OpperWinther2005,Seeger2008,OpperArchambeau2009}. Mean field approaches are more broadly applied but the deteriorating effects of assuming a-posteriori independence have repeatedly been pointed out and discussed in different contexts \citep[e.g.,][]{MacKay2001,IlinValpola2005,TurnerSahani2011a,SheikhEtAl2014,VertesSahani2018}. A further drawback of variational approaches, which they share with most sampling approaches, is the challenge to analytically derive an appropriate approximation for any new generative model. While solutions specific to a generative model are often the best choice w.r.t.\ efficiency, it has been argued \citep[see, e.g.][]{RanganathEtAl2014,SteinrueckenEtAl2019} that the necessary expertise to derive such solutions is significantly taxing the application of variational approaches in practice.

The drawbacks of variational approaches have motivated novel research directions. Early on, mean field approaches were generalized, for instance, to contain mutual dependencies (e.g., structured mean field; \citealt[][]{SaulJordan1996,Bouchard2009,Murphy2012,MacKay2003}) and, more recently, methods to construct arbitrarily complex approximations to posteriors, e.g., using normalizing flows were suggested \citep[][]{RezendeMohamed2015,KingmaEtAl2016}. 
Ideally, a variational approximation should be both very efficiently scalable as well as generally applicable. Also with variational inference becoming increasingly popular
for training deep unsupervised models \citep[e.g.,][]{RezendeEtAl2014,KingmaEtAl2016} the significance of fast and flexible variational methods has further increased.
Not surprisingly, however, increasing both scalability and generality represents a major challenge because in general a trade-off can be observed between the flexibility
of the used variational method on the one hand, and its scalability and task performance on the other.


In order to contribute to novel methods that are both flexible and scalable, we here consider generative models with discrete latents and explore the combination of variational and evolutionary optimization.
For our purposes, we use truncated posteriors as family of variational distributions. In terms of scalability, truncated posteriors suggest themselves as their application has enabled training of very large generative models \citep[][]{SheikhLucke2016,ForsterLucke2018Sublinear}. Furthermore, the family of truncated posteriors is directly defined by the joint distribution of a given generative model, which allows for algorithms that are generally applicable to generative models with discrete latents.
While truncated posteriors have been used and evaluated in a series of previous studies \citep[e.g.][]{LuckeEggert2010,DaiLucke2014,SheltonEtAl2017}, they have not been optimized variationally, i.e., rather than seeking the optimal members of the variational family, truncations were estimated by one-step feed-forward functions. Instead, we here use a
fully variational optimization loop which improves the variational bound by using evolutionary algorithms. For mixture models, fully variational approaches based on truncated posteriors have recently been suggested \citep[][]{ForsterEtAl2018,ForsterLucke2018Sublinear} but they exploit the non-combinatorial nature of mixtures to derive speed-ups for large scales \citep[also compare][]{HughesSudderth2016}. Here we, for the first time, apply a fully variational approach based on truncated posteriors in order to optimize more complex generative models (see \citealt{LuckeEtAl2018} and \citealt{GuiraudEtAl2018} for preliminary results).

\section{Evolutionary Expectation Maximization}
\label{sec:eem}
A probabilistic generative model stochastically generates data points (here referred to as~$\yVec$\,) using a set of hidden (or latent) variables (referred to as~$\sVec$). The generative process can be formally defined by a joint probability $p(\sVec,\yVec \mid \Theta)$, where $\Theta$ is the set of all model parameters. We will introduce concrete models in Section\,\ref{sec:applications}. To start, let us consider general models with binary latents. In such a case $p(\yVec \mid \Theta) = \sum_{\sVec}\, p(\sVec,\yVec\mid\Theta)$ where the sum is taken over all possible configurations of the latent variable $\sVec\in\{0,1\}^H$ with $H$ denoting the length of the latent vector.
Given a set of $N$ data points $\mathcal{Y}=\lbrace\yVecN\rbrace_{n=1,\dots,N}$, we seek parameters $\Theta$ that maximize the data likelihood $\LL(\Theta)=\prod_{n=1}^{N} p(\yVecN \mid \Theta)$.
Here we use an approached based on Expectation Maximization \citep[EM; e.g.][for a review]{GuptaChen2011}. Instead of maximizing the (log-)likelihood directly, we follow, e.g.\ \citeauthor{SaulJordan1996} (\citeyear{SaulJordan1996}) and \citeauthor{NealHinton1998} (\citeyear{NealHinton1998}), and iteratively increase a variational lower bound (referred to as {\em free energy} or {\em ELBO}) which is given by:
\begin{equation}
  \label{eq:free_energy_binary_latent}
  \FF(q,\Theta) = \sum_{n=1}^{N} \sum_{\sVec} \qn(\sVec\,)\,\log \big(p(\yVecN,\sVec \mid \Theta)\big)\, +\,\sum_{n=1}^{N} H[\qn].
\end{equation}
$\qn(\sVec\,)$ are variational distributions and $H[\qn] = - \sum_{\sVec} \qn(\sVec\,)\, \log \big(\qn(\sVec\,)\big)$ denotes the entropy of these distributions. We seek distributions $\qn(\sVec\,)$ that approximate the intractable posterior distributions $p(\sVec \mid \yVecN,\Theta)$ as well as possible and that, at the same time, result in tractable parameter updates. 
%
%
If we denote the parameters of the variational distributions by $\Lambda$, then a variational EM algorithm consists of iteratively maximizing $\mathcal{F}(\Lambda, \Theta)$ \wrt $\Lambda$ in the variational E-step and \wrt $\Theta$ in the M-step. In this respect, the M-step can maintain the same functional form as for exact EM but expectation values now have to be computed with respect to the variational distributions. 

\subsection{Evolutionary Optimization of Truncated Posteriors}
\label{subsec:evolutionary_optimization}
Instead of using specific functional forms such as Gaussians or factored (mean field) distributions for $\qn(\sVec\,)$, we choose, for our purposes, truncated posterior distributions \citep[see, e.g.,][]{LuckeEggert2010,SheikhEtAl2014,SheltonEtAl2017,LuckeForster2019}:
\begin{equation}
  \label{eq:tvem_truncated_posterior}
  q^{(n)}(\sVec \mid \KKn, \ThetaHat) \defeq \frac{p \left(\sVec \mid \yVecN, \ThetaHat \right) }{\sum\limits_{\sVec^{\,\prime} \in \KKn} p \left(\sVec^{\,\prime} \mid \yVecN, \ThetaHat \right)}\delta ( \sVec \in \KKn ) = \frac{p \left(\sVec , \yVecN \mid \ThetaHat \right) }{\sum\limits_{\sVec^{\,\prime} \in \KKn} p \left(\sVec^{\,\prime} , \yVecN \mid \ThetaHat \right)}\delta ( \sVec \in \KKn ),
\end{equation}
where $\delta ( \sVec \in \KKn )$ is equal to 1 if a state $\sVec$ is contained in the set $\KKn$ and 0 otherwise.
The variational parameters are now given by $\Lambda=(\KK,\ThetaHat)$ where $\KK=\lbrace\KKn\rbrace_{n=1,\dots,N}$. With this choice of variational distributions, expectations \wrt the full posterior can be approximated by efficiently computable expectations \wrt truncated posteriors \eqref{eq:tvem_truncated_posterior}:
\begin{equation}
  \label{eq:tvem_truncated_expectations}
  \left<g(\sVec\,)\right>_{q^{(n)}} = \frac{\sum\limits_{\sVec \in \cK^{(n)}}g(\sVec\,)\ p(\sVec, \yVecN \mid \ThetaHat)}{\sum\limits_{\sVec^{\,\prime} \in \cK^{(n)}} p(\sVec^{\,\prime}, \yVecN \mid \ThetaHat)}
\end{equation}
Instead of estimating the relevant states of $\KKn$ using feedforward (preselection) functions \citep[e.g.][]{LuckeEggert2010,SheikhEtAl2014,SheikhEtAl2019}, we here apply a fully variational approach, i.e., we define a variational loop to optimize the variational parameters. Based on the specific functional form of truncated posteriors, optimal variational parameters $\Lambda=(\KK,\ThetaHat)$ are given by setting $\ThetaHat=\Theta$ and by seeking $\KK$ which optimize \citep[see][for details]{Lucke2019}:
\begin{equation}
  \label{eq:tvem_free_energy}
  \mathcal{F}(\cK^{(1\ldots N)}, \Theta) = \sum_{n=1}^{N} \log\Big( \sum_{\sVec \in \cK^{(n)}} \genmodeln \Big).
\end{equation}
Equation\,\ref{eq:tvem_free_energy} represents a reformulation of the variational lower bound \eqref{eq:free_energy_binary_latent} for $\ThetaHat=\Theta$. Because of the specific functional form of truncated posteriors, this reformulation does not contain an explicit entropy term, which allows for an efficient optimization using pairwise comparisons. 
More concretely, the variational bound is provably increased in the variational E-step if the sets $\KKn$ are updated by replacing a state $\sVec$ in $\KKn$ with a new state $\sVecNew$ such that:
\begin{equation}
  \label{eq:tvem_criterion}
   p(\sVecNew,\yVecN \mid \Theta) \,\, > \,\, p(\sVec, \yVecN \mid \Theta)\,,
\end{equation}
where we make sure that any new state $\sVecNew$ is not already contained in $\KKn$. By successively replacing states (or bunches of states), we can keep the size of each set $\KKn$ constant. We use the same constant size $S$ for all sets (i.e.\ $|\KKn|=S$ for all $n$), which makes $S$ an important parameter for the approximation.
The crucial remaining question is how new states can be found such that the lower bound \eqref{eq:tvem_free_energy} is increased efficiently. 

A distinguishing feature when using the family of truncated posteriors as variational distributions is the type of variational parameters, which are given by sets of latent states (i.e.\ by the sets $\KKn$). 
As these states, for binary latents, are given by bit vectors ($\sVec\in\{0,1\}^H$), we can here interpret them as genomes of individuals. Evolutionary algorithms (EAs) then emerge as a very natural choice: we can use EAs to mutate and select the bit vectors $\sVec\in\{0,1\}^H$ of the sets $\KKn$ in order to maximize the lower bound \eqref{eq:tvem_free_energy}. In the EA terminology, the variational parameters $\KKn$ then
become {\em populations} of individuals, where each {\em individual} is defined by its latent state $\sVec\in\{0,1\}^H$ (we will in the following use {\em individual}
to also refer to its genome/latent state).

\begin{figure}
  \centering
  \vspace{-1cm}
  \includegraphics[width=\linewidth]{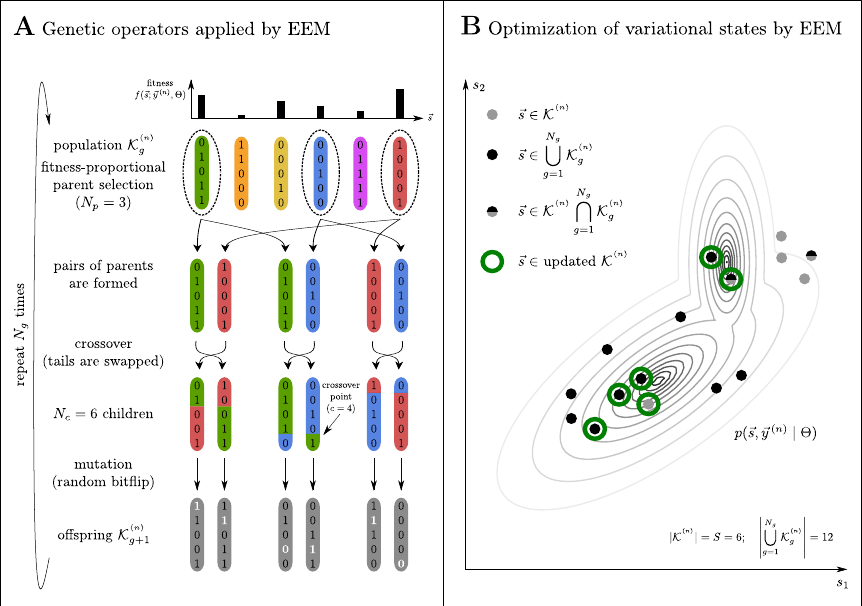}
  \caption{Illustration of the variational optimization using genetic operators.}
\label{fig:eem_illustration}
\end{figure}

For each population $\KKn$, we will use EAs with standard genetic operators (parent selection, mutation and crossover) in order to produce offspring, and we will then use the offspring in order to improve each population. The central function for an EA is the fitness function it seeks to optimize. For our purposes, we will define
the fitness $\mathit{f}(\sVec;\yVecN,\Theta)$ of the individuals $\sVec$ to be a monotonic function of the joint $p(\sVec,\yVecN \mid \Theta)$ of the given generative model, i.e., we define the fitness function to satisfy:
\begin{equation} 
  \label{eq:eem_joint_and_fitness}
  \mathit{f}(\sVecNew;\yVecN,\Theta) \,\, > \,\, \mathit{f}(\sVec; \yVecN,\Theta) \;\;\;\; \Leftrightarrow \;\;\;\; p(\sVecNew,\yVecN \mid \Theta) \,\, > \,\, p(\sVec,\yVecN \mid \Theta)\,.
\end{equation}
A fitness satisfying (\ref{eq:eem_joint_and_fitness}) will enable the selection of parents that are likely to produce offspring with high
joint probability $p(\sVecNew,\yVecN \mid \Theta)$. Before we detail how the offspring is used to improve a population $\KKn$, we first
describe how the EA generates offspring.
For each population $\KKn$, we set $\KKn_{0}=\KKn$ and then iteratively generate new generations $\KKn_{g}$ by successive application of Parent Selection, Crossover, and Mutation operators:
%
%
\\
\\
\noindent{\it Parent Selection.} To generate offspring, $N_p$ parent states are first selected from the initial population $\KKn_{0}$ (see Figure\,\ref{fig:eem_illustration}). Ideally, the parent selection procedure should be balanced between exploitation of parents with high fitness (which might be expected to be more likely to produce children with high fitness) and exploration of mutations of poor performing parents (which might be expected to eventually produce children with high fitness while increasing population diversity). Diversity is crucial, as the $\KKn$ are sets of unique individuals and therefore the improvement of the overall fitness of the population depends on generating as many as possible {\em different} children with high fitness.
In our numerical experiments, we explore both random uniform selection of parents and fitness-proportional selection of parents (i.e., parents are selected with a probability proportional to their fitness). 
For the latter case, we here define the fitness as follows:
%
\begin{equation}
  \label{eq:eem_fitness}
  \mathit{f}(\sVec;\yVecN,\Theta) \defeq \widetilde{\log p}(\sVec,\yVecN\mid\Theta) + \mathrm{const}.
\end{equation}
The term $\widetilde{\log p}(\sVec,\yVecN\mid\Theta)$ denotes a monotonously increasing function of the 
joint $p(\sVec,\yVecN\mid\Theta)$ which is more efficiently computable and which has better numerical stability than the joint itself.  $\widetilde{\log p}(\sVec,\yVecN\mid\Theta)$ is defined as the logarithm of the joint where summands that do not depend on $\sVec$ have been elided (see Appendix\,\ref{app:lpj} for examples). 
As the fitness \eqref{eq:eem_fitness} satisfies \eqref{eq:eem_joint_and_fitness}, the parents are likely to have high joint probabilities if they are selected proportionally to $\mathit{f}(\sVec;\yVecN,\Theta)$.
The constant term in \eqref{eq:eem_fitness} is introduced to ensure that the fitness function always takes positive values for fitness proportional selection.
The constant is defined to be $\mathrm{const} = \big| 2\min_{\sVec\in\KKn_{g}}\widetilde{\log p}(\sVec,\yVecN\mid\Theta)\big|$. For a given population of individuals $\KKn_{g}$, the offset $\mathrm{const}$ is thus constant throughout the generation of the next population $\KKn_{g+1}$. 
According to \eqref{eq:eem_fitness} the fitness function can be evaluated efficiently given that the joint $p(\sVec,\yVecN\mid\Theta)$ can efficiently be computed (which we will assume here).
\\
\\
\noindent{\it Crossover.} For the crossover step, all the parent states are paired in each possible way. Each pair is then assigned a number $c$ from $1$ to $H-1$ with uniform probability ($c$ defines the single crossover point). Finally each pair swaps the last $H-c$ bits to generate offspring. The procedure generates $N_c=N_p(N_p-1)$ children. The crossover step can be skipped, making the EA more lightweight but decreasing variety in the offspring.
\\
\\
\noindent{\it Mutation.} Finally, each child undergoes one random bitflip to further increase offspring diversity. In our experiments, we compare results of random uniform selection of the bits to flips with a more refined sparsity-driven bitflip algorithm (the latter scheme assigns to 0's and 1's different probabilities of being flipped in order to produce children with a sparsity compatible with the one learned by a given model; details in Appendix\,\ref{app:eem_sparseflips}).
If the crossover step is skipped, the parent states are copied $N_m$ times ($N_m$ can be chosen between $1\leq N_m \leq H$) and offspring is produced by mutating each copy by one random bitflip, producing $N_c=N_p N_m$ children.
\\
\\
\noindent
The application of the genetic operators is iteratively repeated. The offspring $\KKn_{g+1}$ produced by the population $\KKn_{g}$ becomes the new population from which parents are selected (Figure\,\ref{fig:eem_illustration}). After $N_g$ iterations, we obtain a large number of new individuals (i.e., all the newly generated populations $\KKn_{g}$ with $g=1,\ldots,N_g$). The new populations of individuals can now be used to improve the original population $\KKn$ such that a higher value for the lower bound \eqref{eq:tvem_free_energy} is obtained. To find the best new population for a given $n$, we collect all generated populations (all $\KKn_{g}$ with $g=1,\ldots,N_g$) and unite them with $\KKn$. We then reduce this larger set back to the population size $S$, and we do so by only keeping those $S$ individuals with the highest\footnote{Instead of selecting the $S$ elements with largest joints, we in practice remove the $(|\KKn|-S)$ elements of $\KKn$ with the lowest joint probabilities. Both procedures are equivalent.} joints $p(\sVec,\yVecN \mid \Theta)$. 
By only keeping the states/individuals with highest joints, the update procedure for $\KKn$ is guaranteed to monotonically increase the variational lower bound \eqref{eq:tvem_free_energy} since the variational bound is increased if criterion \eqref{eq:tvem_criterion} is satisfied.
%

\begin{algorithm}[t]
\label{alg:eem}
\vspace{1.5mm}
define selection, crossover and mutation operators; \\ 
set hyperparameters $S$, $N_g$, etc.;\\ 
initialize model parameters $\Theta$;\\ 
for each $n$: populate set $\KKn$ with $S$ latent states ($|\KKn|=S$);\\ 
\Repeat{parameters $\Theta$ have sufficiently converged}
{  
  \phantom{x=x}\ \\[-2mm]
  \For{$n=1,\ldots,N$\vspace{1.5mm}}
  {        
    set $\KKn_0 = \KKn$;\algBreak  
    \For{$g=1,\ldots,N_g$\vspace{1.5mm}}
    {      
      $\KKn_{g} = \text{mutation}\left(\text{crossover}\left(\text{selection} \left(\KKn_{g-1}\right)\right)\right)$; \algBreak
      $\KKn\,=\,\KKn \cup \KKn_g$;\algBreak
    }   
    remove those $(|\KKn|-S)$ elements $\sVec$ in $\KKn$ with lowest $p(\sVec,\yVecN\,|\,\Theta)$;\algBreak  
  }
  use M-steps with \eqref{eq:tvem_truncated_expectations} to update $\Theta$;\algBreak
  }
\caption{Evolutionary Expectation Maximization (EEM).}
\end{algorithm}
Algorithm~\ref{alg:eem} summarizes the complete variational algorithm; an illustration is provided in Figure\,\ref{fig:eem_illustration}. As the variational E-step is realized using an evolutionary algorithm, Algorithm~\ref{alg:eem} will from now on be referred to as {\em Evolutionary Expectation Maximization} (EEM).  The EEM algorithm can be trivially parallelized over data-points. For later numerical experiments, we will use a parallelized implementation of EEM that can be executed on several hundreds of computing cores (for further technical details see Appendix\,\ref{app:numerical_experiments_tech} and also compare \citealt[][]{SalimansEtAl2017}). For the numerical experiments, we will indicate which parent selection procedure (“fitparents” for fitness-proportional selection, “randparents” for random uniform selection) and which bitflip algorithm (“sparseflips” or “randflips”) are used in the variational loop. The term “cross” will be added to the name of the EA when crossover is employed.

\subsection{Related Work on Applications of Evolutionary Approaches to Learning}
Evolutionary algorithms (EAs) have repeatedly been applied to learning algorithms \citep[e.g.][]{Goldberg2006}. In the context of reinforcement learning, for instance, evolution strategies have successfully been used as an optimization alternative to Q-learning and policy gradients \citep{SalimansEtAl2017}.
EAs have also previously been applied to unsupervised learning based on generative models, also in conjunction with (or as substitute for) expectation maximization (EM).
\citet{MyersEtAl1999} used EAs to learn the structue of Bayes Nets, for instance. \citet{PernkopfBouchaffra2005} have used EAs for clustering based on Gaussian mixture models (GMMs), where GMM parameters are updated relatively conventionally using EM. EAs are then used to select the best GMM models for the clustering problem (using a minimum description length criterion). Work by \citep[][]{TohkaEtAl2007} uses EAs for a more elaborate initialization of Finite Mixture Models, which is followed by EM. Work by \citet[][]{TianEtAl2011} goes further and combines EAs with a variational Bayes approach for GMMs. The algorithm they suggested is based on a fully Bayesian GMM approach with hyper-priors in the place of standard GMM parameters. The algorithm then first uses an EA to find good hyperparameters; given the hyperparameters the GMM is then trained using standard (but variational) EM iterations. The use of EAs by \citet{PernkopfBouchaffra2005} and \citet[][]{TianEtAl2011} is similar to their use for deep neural network (DNN) training.
For DNNs, EAs are applied to optimize network hyperparameters in an outer loop \citep[][etc.]{StanleyMiikkulainen2002,LoshchilovHutter2016,RealEtAl2017,SuganumaEtAl2017,OehmckeAndKramer2018} while DNN weight parameters are at the same time trained using standard back-propagation algorithms. Yet other approaches have applied EAs to directly optimize a clustering objective, using EAs to {\em replace} EM approaches for optimization \citep[compare][]{HruschkaEtAl2009}. Similarly, in a non-probabilistic setting, EAs have been used to replace back-propagation for DNN training \citep[see, e.g.,][and references therein]{PrellbergAndKramer2018}. Also in a non-probabilistic setting, for sparse coding with weights optimized based on a $l_1$-penalized reconstruction objective \citep[e.g.][]{OlshausenField1996}, an EA tailored to this objective has been used to address maximum a-posteriori optimization \citep[][]{AhmadiAndSalari2016}.
%
In contrast to all such previous applications, the EEM approach (Algorithm~\ref{alg:eem}) applies EAs as an integral part of variational EM, i.e.,  
EAs address the key optimization problem of variational EM in the inner variational loop.

\subsection{Data Estimator}
\label{subsec:estimator}
As part of the numerical experiments, we will apply EEM to fit generative models to corrupted data which we will aim to restore; for instance we will apply EEM to denoise images corrupted by additive noise or to restore images with missing data (see Section\,\ref{sec:performance}). Here we illustrate how an estimator for such data reconstruction can be derived (details are given in Appendix\,\ref{app:data_estimator}). 
Generally speaking, our aim is to compute estimates $\yVecEst$ based on observations $\yVecObs$. 
If data points are corrupted by noise but not by missing values, we define the observations $\yVecObs$ to be equal to the data points $\yVec \in \mathds{R}^D$ and we compute estimates $\yDEst$ for every $d=1,\dots,D$ with $D$ denoting the dimensionality of the data. In the presence of missing values, we define data points to consist of an observed part $\yVecObs$ and a missing part $\yVecMiss = \yVec\,\backslash\yVecObs$ and we aim to compute estimates $\yVecEst$ for the missing part.
A data estimator can be derived based on the posterior predictive distribution $p(\yVecEst\mid\yVecObs)$. Here we will sketch how to derive an estimator for general models with binary latents and continuous observables (in Appendix\,\ref{app:data_estimator} we provide a detailed derivation and show how to extend it to models with binary-continuous latents).
The posterior predictive distribution for a model with binary latents and continuous observables is given by:
\begin{align}
p(\yVecEst\mid \yVecObs, \Theta)
&=\sum_{\sVec} p(\yVecEst\,\mid\,\sVec, \Theta) p(\sVec\,\mid\,\yVecObs, \Theta).
\end{align}
We choose to take expectations \wrt the posterior predictive distribution which leads to the following estimator:
\begin{align}
\label{eq:binlat_general_data_estimator_final}
\langle \yDEst \rangle_{p(\yVecEst\,\mid\,\yVecObs, \Theta)}
&= \left \langle \langle \yDEst \rangle_{p(\yDEst\,\mid\,\sVec, \Theta)} \right \rangle_{p(\sVec\,\mid\,\yVecObs, \Theta)}.
\end{align}
The inner expectation in \eqref{eq:binlat_general_data_estimator_final} can be identified with the mean of the distribution of the observables given the latents. This distribution is part of the definition of the generative model (compare Section\,\ref{subsec:models}). The outer expectation in $\eqref{eq:binlat_general_data_estimator_final}$ is taken \wrt the posterior distribution of the latents. Such expectation can efficiently be approximated using truncated expectations \eqref{eq:tvem_truncated_expectations}. Details and examples of data estimators for concrete models are given in Appendix\,\ref{app:data_estimator}.

\section{Application to Generative Models, Verification and Scalability}
\label{sec:applications}
Considering Algorithm\,\ref{alg:eem}, observe that probabilistic inference for the algorithm is fully defined by the joint $p(\sVec,\yVecN\,|\,\Theta)$ of a given generative model and by a set of hyperparameters for the optimization procedure. No additional and model specific derivations are required for the variational E-step, which suggests Algorithm\,\ref{alg:eem} for ``black box'' inference for models with binary latents. Using three different generative models, we here (A)~verify this ``black box'' applicability, (B)~numerically evaluate the algorithm's ability to recover generating parameters, and (C)~show scalability of the approach by applying it to large-scale generative models. To be able to study the properties of the novel variational optimization procedure, we consider generative models for which parameter update equations (M-steps) have been derived previously.

\subsection{Used Generative Models}
\label{subsec:models}
The generative models we use are noisy-OR Bayes Nets (binary latents and binary observables), Binary Sparse Coding (binary latents and continuous observables), and Spike-and-Slab Sparse Coding (continuous latents and continuous observables).
\\
\\
\noindent{\it Noisy-OR Bayes Nets.} A noisy-OR Bayes Net \citep[e.g.,][]{SingliarHauskrecht2006,JerniteEtAl2013,RotmenschEtAl2017} is a non-linear bipartite data model with all-to-all connectivity between the layer of hidden and observed variables (and no intra-layer connections). Latents and observables both take binary values, i.e.\ $\sVec \in \lbrace 0,1 \rbrace^{H}$ and $\yVec \in\lbrace 0,1 \rbrace^{D}$ with $H$ and $D$ denoting the dimensions of the latent and observed spaces. The model assumes a Bernoulli prior for the latents, and non-zero latents are then combined via the noisy-OR rule:
\begin{gather}
  \label{eq:nor_gen_model}
  \sPrior = \prod_{h=1}^{H} \BBernou(s_h;\pi_h), \;\;\;\;\;\;\;  \pygs = \prod_{d=1}^{D} N_d(\sVec\,)^{y_d} (1-N_d(\sVec\,))^{1-y_d},\quad
\end{gather}
where $\BBernou(s_h;\pi_h) = \pi_h^{s_h}(1-\pi_h)^{1-s_h}$ and where $N_d(\sVec\,) \defeq 1 - \prod_{h=1}^{H}(1 - W_{dh}s_h)$. 
In the context of the noisy-OR model $\Theta = \{ \vec{\pi}, W \}$, where $\vec{\pi}$ is the set of values $\pi_h \in [0,1]$ representing the prior activation probabilities for the hidden variables $s_h$ and $W$ is a $D{\times}H$ matrix of values $\Wdh \in [0,1]$ representing the probability that an active latent variable $s_h$ activates the observable $y_d$. 
M-step update rules for the noisy-OR model can be derived by inserting \eqref{eq:nor_gen_model} into the lower bound \eqref{eq:free_energy_binary_latent} and by then taking derivatives of the resulting expression \wrt all model parameters. The update rule for the $W$ parameter does not allow for a closed-form solution and a fixed-point equation is employed instead. The resulting M-step equations are shown in Appendix\,\ref{app:mstep_equations}.
\\
\\
\noindent{\it Binary Sparse Coding.} Binary Sparse Coding (BSC; \citealp[][]{HaftEtAl2004,HennigesEtAl2010}) is an elementary sparse coding model for continuous data with binary latent and continuous observed variables. Similar to standard sparse coding \citep[][]{OlshausenField1997,LeeEtAl2007} BSC assumes that given the latents, the observables follow a Gaussian distribution. BSC and standard sparse coding differ from each other in their latent variables. In standard sparse coding, latents take continuous values and they are typically modeled e.g. with Laplace or Cauchy distributions. BSC uses binary latents $\sVec \in \lbrace 0,1 \rbrace^{H}$ which are assumed to follow a Bernoulli distribution $\BBernou(s_h;\pi)$ with the same activation probability $\pi$ for each hidden unit $s_h$. The combination of the latents is described by a linear superposition rule. Given the latents, the observables $\yVec \in \RRR^{D}$ are independently and identically drawn from a Gaussian distribution:
\begin{gather}
  \label{eq:bsc_gen_model}
  \sPrior = \prod_{h=1}^H \BBernou(s_h;\pi), \;\;\;\;\;\;\; \pygs = \prod_{d=1}^D \NGauss(y_d;\sum_{h=1}^{H} W_{dh}s_h,\sig^2)
\end{gather}
The parameters of the BSC model are $\Theta=(\pi,\sig,W)$. $W$ is a $D\times H$ matrix whose columns contain the weights associated with each hidden unit; $\sig$ determines the standard deviation of the Gaussian. 
The M-step equations for BSC can be derived analogously to the noisy-OR model by inserting \eqref{eq:bsc_gen_model} into \eqref{eq:free_energy_binary_latent} and by then optimizing the resulting expression \wrt each model parameter \citep[compare, e.g.,][]{HennigesEtAl2010}. As opposed to noisy-OR, each of the BSC update equations can be derived in closed-form. The explicit expressions are shown in Appendix\,\ref{app:mstep_equations}.
\\
\\
\noindent{\it Spike-and-Slab Sparse Coding.} As last example we consider a more expressive data model and use EEM to optimize the parameters of a spike-and-slab sparse coding (SSSC) model. SSSC extends the generative model of BSC in the sense that it uses a spike-and-slab instead of a Bernoulli prior distribution. The SSSC model has been used in a number of previous studies and in a number of variants \citep[][and many more]{TitsiasLazaro2011,LuckeSheikh2012,GoodfellowEtAl2012,SheikhEtAl2014}. It can be formulated using two sets of latents, $\sVec \in \lbrace 0,1 \rbrace^{H}$ and $\zVec \in \RRR^{H}$, which are combined via pointwise multiplication s.t.\ $\szVecBr_{h}=s_{h}z_{h}$. Here we take the continuous latents to be distributed according to a multivariate Gaussian with mean $\muVec$ and a full covariance matrix $\Psi$. The binary latents follow a Bernoulli distribution with individual activation probabilities $\pi_h$, $h=1,\dots,H$:
\begin{equation}
  \label{eq:sssc_gen_model_prior}
  \sPrior = \prod_{h=1}^H \BBernou(s_h;\pi_h), \;\;\;\;\;\;\; \pzgt = \NGauss(\zVec; \muVec,\Psi). \\  
\end{equation}
As in standard sparse coding, SSSC assumes components to combine linearly. The linear combination then determines the mean of a univariate Gaussian for the observables:
\begin{equation}
  \label{eq:sssc_gen_model_conditional} 
  \pygsz = \NGauss(\yVec;\,\sum_{h=1}^{H} \vec{W}_h s_{h}z_{h},\sigma^2\eye)
\end{equation}
where $W$ is a $D\times H$ matrix with columns $\vec{W}_h$ and where $\eye$ is the unit matrix. The M-step update rules of the SSSC model can be derived by taking derivatives of the free energy
\begin{equation}
  \label{eq:free_energy_binary_continuous_latent}
\FF(q,\Theta) = \sum_{n=1}^{N} \bigl< \log p(\yVecN,\sVec,\zVec \mid \Theta)\bigr>_{\qn} \, +\,\sum_{n=1}^{N} H[\qn]
\end{equation}
\wrt each of the model parameters $\Theta=(\piVec,\sigma,W,\muVec,\Psi)$.
The term $\bigl<f(\sVec,\zVec)\bigr>_{\qn}$ in \eqref{eq:free_energy_binary_continuous_latent} denotes the expectation of $f(\sVec,\zVec)$ \wrt a variational distribution $\qn(\sVec,\zVec)$:
\begin{equation}
  \label{eq:sssc_exp_values}
  \bigl<f(\sVec,\zVec)\bigr>_{\qn} = \sum_{\sVec} \int \qn(\sVec,\zVec) f(\sVec,\zVec)\,\dz.
\end{equation}
The term $H[\qn]$ in \eqref{eq:free_energy_binary_continuous_latent} denotes the entropy $-\bigl<\log \qn(\sVec,\zVec)\bigr>_{\qn}$ of the variational distributions. 
For a detailed derivation of the SSSC M-step equations see \citet[][]{SheikhEtAl2014} and compare \citet[][]{GoodfellowEtAl2012}.
For SSSC, all M-step updates (A)~have closed-form solutions and (B)~contain as arguments expectation values \eqref{eq:sssc_exp_values}. Importantly for applying EEM, all these expectation values can be reformulated as expectations \wrt the posterior over the binary latent space. For more details see Appendix\,\ref{app:mstep_equations} and compare \citet[][]{SheikhEtAl2014}. Based on the reformulations, the expectations \wrt the posterior over the binary latent space can be approximated using the truncated expectations \eqref{eq:tvem_truncated_expectations}. 
Since the joint $p(\sVec,\yVecN \mid \Theta)$ of the SSSC model is computationally tractable and given by
\begin{gather} \label{eq:_sssc_joint_s_y}
  \genmodeln = \BBernou(\sVec;\piVec) \,\, \NGauss(\yVecN;\,\tWs\, \muVec,\Cs),
\end{gather}
with $\Cs = \sigma^2\eye + \tWs\,\Psi\,\tWs\transp$, $(\tWs)_{dh} = W_{dh}s_{h}$ s.t. $W\szVecBr = \tWs\,\zVec$, the variational lower bound \eqref{eq:tvem_free_energy} can be efficiently optimized using Algorithm\,\ref{alg:eem}. The EEM algorithm consequently provides a novel, evolutionary approach to efficiently optimize the parameters of the SSSC data model.
%
%

\subsection{Verification: Recovery of Generating Parameters}
\label{subsec:verification}
First we verified EEM by training noisy-OR, BSC and SSSC models on artificial data for which the ground-truth generative parameters were known. We used the bars test as a standard setup for such purposes \citep[][]{Foeldiak1990,Hoyer2003,LuckeSahani2008}.
We started with a standard bars test using a weight matrix $W^\gen$ whose $H^\gen=2\sqrt{D}$ columns represented generative fields in the form of horizontal and vertical bars. We considered a dictionary of $H^\gen=10$ \citep[][]{LuckeSahani2008,LuckeEggert2010,LuckeEtAl2018,SheikhLucke2016} components. For noisy-OR, we set the amplitudes of the bars and of the background to a value of 0.8 and 0.1, respectively; for BSC and SSSC the bar amplitudes were uniformly randomly set to 5 and -5 and the background was set to 0. For BSC and SSSC, we chose $(\sigma^\gen)^2=1$ as ground-truth variance parameter; for SSSC we furthermore defined $\muVec^\gen = \vec{0}$ and $\Psi^\gen=\eye$. 

\begin{figure}
  \centering
  \input{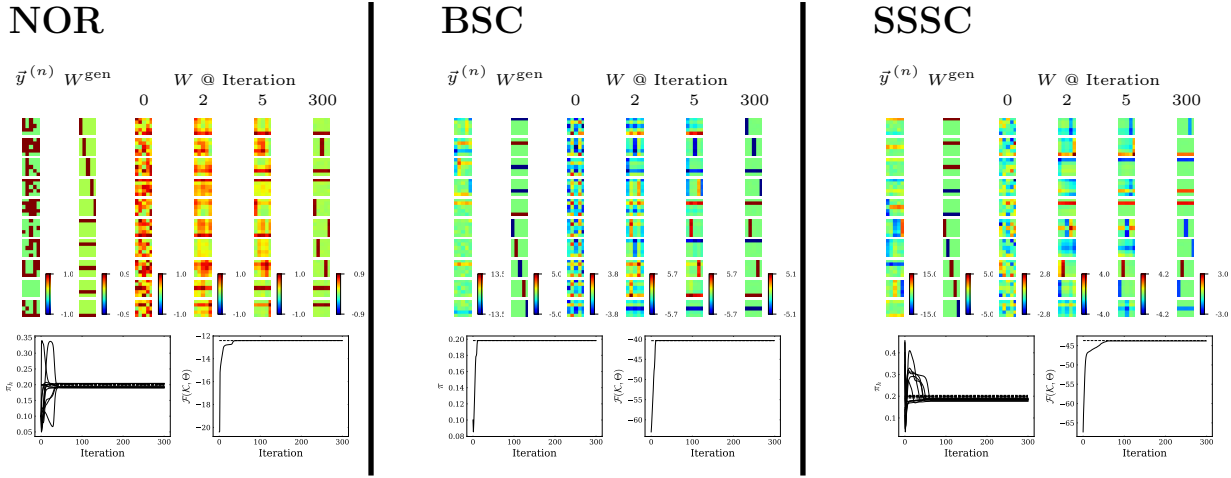}
  \vspace{-.1in}
  \caption{Training noisy-OR (NOR), BSC and SSSC models on artificial data using EEM.}
  \label{fig:norbscssscBars}
\end{figure}

For each model, 30 datasets of $N=5,000$ samples were generated. We used $\pi_h^\gen = \frac{2}{H^\gen}\,\forall\,h=1,\dots,H^\gen$ (see Figure~\ref{fig:norbscssscBars} for a few examples). We then applied EEM to train noisy-OR, BSC and SSSC models with $H=H^\gen=10$ components (EEM hyperparameters are listed in Table\,\ref{tab:hyperparams} in Appendix\,\ref{app:numerical_experiments_tech}).
For noisy-OR, priors $\pi_h$ were initialized to an arbitrary sparse value (typically $\frac{1}{H}$) while the initial weights $W^\init$ were sampled from the standard uniform distribution. The $\KKn$ sets were initialized by sampling bits from a Bernoulli distribution that encouraged sparsity. In our experiments, the mean of the distribution was set to $\frac{1}{H}$.
For BSC, the initial value of the prior was defined $\pi^\init=\frac{1}{H}$; the value of $(\sigma^\init)^2$ was set to the variance of the data points averaged over the observed dimensions. The columns of the $W^\init$ matrix were initialized with the mean of the data points to which some Gaussian noise with a standard deviation of $0.25\,\sigma^\init$ was added. The initial values of the $\KKn$ sets were drawn from a Bernoulli distribution with $p(s_h=1)=\frac{1}{H}$. To initialize the SSSC model, we uniformly randomly drew $\piVec^\init$ and $\muVec^\init$ from the interval $[0.1,0.5]$ and $[1,5]$, respectively \citep[compare][]{SheikhEtAl2014} and set $\Psi^\init$ to the unit matrix. For SSSC, we proceeded as we had done for the BSC model to initialize the dictionary $W$, the variance parameter $\sigma^2$ and the sets $\KKn$. The evolution of the model parameters during learning is illustrated in Figure~\ref{fig:norbscssscBars} for an exemplary run of the ``fitparents-randflip'' EA (illustrations of the parameters $\sigma$ of the BSC model and $\Psi$, $\muVec$ and $\sigma$ of the SSSC model are depicted in Figure\,\ref{fig:app_bscssscBars} in Appendix\,\ref{app:numerical_results}).

Figure~\ref{fig:norbscssscBars} illustrates that the ground-truth generative parameters can accurately be recovered for each model using EEM. The weights, latent means and latent covariances of the SSSC model are recovered in scaled versions \wrt the ground-truth values. This effect can be considered as a consequence of the spike-and-slab prior which allows for multiple equivalent solutions for the amplitudes of $W$ and $\zVec$ given a data point.
We further used the setup with artificial data and known ground-truth generative parameters to explore EAs with different types of combinations of genetic operators. Different combinations of genetic operators resulted in different degrees of ability to recover the ground-truth parameters. For instance, for noisy-OR trained with EEM, the combination of random uniform parent selection, single-point crossover and sparsity-driven bitflip (``randparents-cross-sparseflips'') resulted in the best behavior for bars test data \citep[also compare][]{GuiraudEtAl2018}. For BSC and SSSC trained with EEM, the best operator combinations were ``fitparents-cross-sparseflips'' and ``fitparents-randflips'', respectively (compare Figure\ \ref{fig:app_norbscssscReliability} in Appendix\,\ref{app:numerical_results}). 
In general, the best combination of operators will depend on the data model and the data set used.

\subsection{Scalability to Large Generative Models}
\label{subsec:scalability}
Having verified the ability of EEM to recover ground-truth generating parameters for the three generative models of Section\,\ref{subsec:models}, we continued with investigating how well the evolutionary approximation was able to optimized large-scale models. Scalability is a property of crucial importance for any approximation method, and large-scale applicability is essential to accomplish many important data processing tasks based on generative models. To investigate scalability, we used natural image patches which are known to be richly structured and which require large models to appropriately model their distribution. We used image patches extracted from a standard image database \citep[][]{HaterenSchaaf1998}. We trained the noisy-OR, BSC and SSSC data models using the ``fitparents-cross-sparseflips'' variant of EEM as we observed this operator combination to perform well across all models for this data.

\begin{figure}
  \centering
  \includegraphics{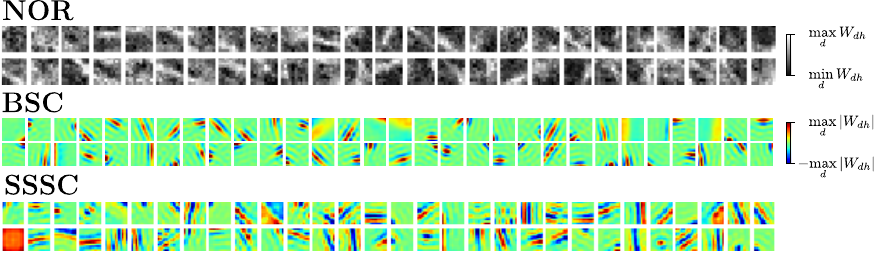}
  \caption{Dictionary elements learned by noisy-OR (top), BSC (middle) and SSSC (bottom) models on natural image patches. The models were trained on separate datasets that had been obtained using different preprocessing schemes (see text). For noisy-OR, a random selection of 50 out of 100 fields is displayed; for BSC the fields corresponding to the 60 out of 300 most active hidden units are shown; for SSSC a random selection of 60 out of 512 fields is displayed. The full dictionaries learned by the individual models are depicted in Figure\,\ref{fig:app_norbscssscHateren} in Appendix\,\ref{app:numerical_results}.}
  \label{fig:norbscssscHateren}
\end{figure}

For noisy-OR, we considered raw images patches, i.e., images which were not mean-free or whitened and reflected light intensities relatively directly. We used image patches that were generated by extracting random $10 \times 10$ subsections of a single $255\times255$ image of overlapping grass wires (part of image 2338 of the database). We clamped the brightest 1\,\% pixels from the dataset, and scaled each patch to have gray-scale values in the range $[0,1]$. From these patches, we then created $N=30,000$ data points $\yVecN$ with binary entries by repeatedly choosing a random gray-scale image and sampling binary pixels from a Bernoulli distribution with parameter equal to the pixel values of the patches with $[0,1]$ gray-scale.
Because of mutually occluding grass wires in the original image, the generating components of the binary data could be expected to follow a non-linear superimposition, which motivated the application of a non-linear generative model such as noisy-OR. For the data with $D=10 \times 10=100$ observables, we applied a noisy-OR model with $H=100$ latents (EEM hyperparameters are listed in Table\,\ref{tab:hyperparams} in Appendix\,\ref{app:numerical_experiments_tech}). During training, we clamped the priors $\pi_h$ to a minimum value of $1 / H$ to encourage the model to make use of all generative fields. Figure\,\ref{fig:norbscssscHateren}\,(top) shows a random selection of 50 generative fields learned by the noisy-OR model (the full dictionary is displayed in Figure\,\ref{fig:app_norbscssscHateren} in Appendix\,\ref{app:numerical_results}). 
Model parameters initialization followed the same procedure as for the bars tests. EEM parameters are listed in Table\,\ref{tab:hyperparams} in Appendix\,\ref{app:numerical_experiments_tech}.
As can be observed, EEM is efficient for large-scale noisy-OR on real data. Many of the generative fields of Figure\,\ref{fig:norbscssscHateren}\,(top) resemble curved edges, which is in line with expectations and with results, e.g., as obtained by \cite{LuckeSahani2008} with another non-linear generative model.

For the BSC generative model, we are not restricted to positive data. Therefore, we can use a whitening procedure as is customary for sparse coding approaches \citep[][]{OlshausenField1997}. We employed $N=100,000$ image patches of size $D=16\times 16=256$ which were randomly picked from the van Hateren dataset. The highest 2\,\% of the pixel values were clamped to compensate for light reflections, and patches without significant structure were removed to prevent amplifications of very small intensity differences. The whitening procedure we subsequently applied is based on ZCA whitening \citep[][]{BellSejnowski1997} retaining 95\,\% of the variance \citep[compare][]{ExarchakisLucke2017}. We then applied EEM to fit a BSC model with $H=300$ components to the data (EEM hyperparameters are listed in Table\,\ref{tab:hyperparams} in Appendix\,\ref{app:numerical_experiments_tech}). To initialize the model and the variational parameters, we proceeded as we had done for the bars test (Section\,\ref{subsec:verification}). 
Figure \ref{fig:norbscssscHateren}\,(middle) depicts some of the generative fields learned by the BSC model. The fields correspond to the 60 hidden units with highest prior activation probability (for the full dictionary see Figure\,\ref{fig:app_norbscssscHateren} in Appendix\,\ref{app:numerical_results}). The obtained generative fields primarily take the form of the well-known Gabor functions with different locations, orientations, phase, and spatial frequencies.

We collected another $N=100,000$ image patches of size $D=12\times12=144$ from the van Hateren dataset to fit a SSSC model with $H=512$ components (EEM hyperparameters listed in Table\,\ref{tab:hyperparams} in Appendix\,\ref{app:numerical_experiments_tech}). We applied the same preprocessing as for the BSC model and initialized the model and the variational parameters similarly to the bars test. 
Figure\,\ref{fig:norbscssscHateren}\,(bottom) shows a random selection of 60 out of the 512 generative fields learned by the SSSC model (the full dictionary is displayed in Figure\,\ref{fig:app_norbscssscHateren} in Appendix\,\ref{app:numerical_results}). 
The experiment shows that EEM can be applied to train complex generative models on large-scale realistic data sets; the structures of the learned generative fields (a variety of Gabor-like and a few globular fields) are in line with observations made in previous studies which applied SSSC models to whitened image patches \citep[see, e.g.,][]{SheikhLucke2016}.

\section{Performance Comparison on Denoising and Inpainting Benchmarks}
\label{sec:performance}
So far, we have verified that EEM can be applied to the training of elementary generative models such as noisy-OR Bayes Nets and Binary Sparse Coding (BSC) but also to the more expressive model of spike-and-slab sparse coding (SSSC) which features a more flexible prior than BSC or standard sparse coding. To perform variational optimization for these models, no additional analytical steps were required and standard settings of optimization parameters were observed to work well. However, an important question is how well EEM is able to optimize model parameters compared to other methods. And more generally, how well do generative model algorithms perform in benchmarks if their parameters are optimized by EEM? Here we will address these questions using two standard and popular benchmarks: image denoising and image inpainting. These two benchmarks offer themselves for performance evaluation, as benchmark data is available for a large range of different algorithms (including algorithms based on generative models among many others).
The data for standard denoising and inpainting benchmarks is continuous. We will consequently focus on the BSC and SSSC generative models. In the following, we will refer to EEM applied to BSC as {\em Evolutionary Binary Sparse Coding} (EBSC) and to EEM applied to SSSC as {\em Evolutionary Spike-and-Slab Sparse Coding} (ES3C). 
As genetic operator combination for the EA we will use ``fitparents-randflips'' for all the following experiments. 
The main reason to prefer this combination over combinations that also use cross-over is computational efficiency (which is a major limiting factor for the following experiments).

%
%
%
%
%
%

\subsection{Algorithm Types and Comparison Categories}
\label{subsec:other_ir_algos}
We will use standard benchmarks to compare EBSC and ES3C \wrt (A)~algorithms based on generative models equal or similar to the ones used in this work, (B)~other probabilistic methods including mixture models and Markov random fields approaches, (C)~non-local image specific methods and (D)~deep neural networks. Table\,\ref{tab:other_ir_algos} gives an overview of the methods we compare to, sorted by algorithm type. To the knowledge of the authors, the listed papers represent the best performing approaches while we additionally included standard baselines such as BM3D, EPLL or MLP (see Table\,\ref{tab:other_ir_algos} for references).

The algorithms listed in Table\,\ref{tab:other_ir_algos} have different requirements that have to be fulfilled for them to be applicable.
To facilitate comparison, we have grouped the benchmarked algorithms into the different categories of prior knowledge they require (see Table\,\ref{tab:requirements}
and Appendix\,\ref{app:discussion_on_evaluation} for a related discussion).
%
As can be observed, some algorithms (e.g., KSVD or BM3D) require the noise level to be known a-priori, for instance. Others do require (often large amounts of)
clean images for training, which typically applies for feed-forward deep neural networks (DNNs). If an algorithm is able to learn without information other than
the corrupted image itself, it is commonly referred to as a ``zero-shot'' learning algorithm \citep[compare, e.g.,][]{ShocherEtAl2018,ImamuraEtAl2019}. Algorithms of the ``zero-shot'' category we compare to are, e.g., MTMKL, BPFA and DIP.
The EEM-based algorithms EBSC and ES3C also fall into the ``zero-shot'' category. In Table\,\ref{tab:requirements}, we have labeled the categories DE1-DE6 for denoising and
IN1-IN6 for inpainting. DE2 could also be referred to as ``zero-shot + noise level'' as the algorithms of that category also do not need additional images. 

\begin{table}
\centering
\newcommand\spac{-0.1cm}
\renewcommand{\aboverulesep}{0ex}
\renewcommand{\belowrulesep}{0ex}
\newcolumntype{K}[1]{>{\centering\arraybackslash}p{#1}}
\resizebox{\linewidth}{!}{
\begin{tabular}[c c c]{|c c c|}
\toprule
\multirow{2}{*}{Acronym}&\multirow{2}{*}{Algorithm Name / Type}&\multirow{2}{*}{Reference}\\
&&\\
\bottomrule
\multicolumn{3}{c}{\vspace{\spac}}\\
\midrule
\multicolumn{3}{|c|}{Sparse Coding / Dictionary Learning}\\
\midrule
MTMKL&Spike-and-Slab Multi-Task and Multiple Kernel Learning&\citealt{TitsiasLazaro2011}\\
BPFA&Beta Process Factor Analysis&\citealt{ZhouEtAl2012}\\
GSC&Gaussian Sparse Coding&\citealt{SheikhEtAl2014}\\
KSVD&Sparse and Redundant Representations Over Trained Dictionaries&\citealt{EladAndAharon2006}\\
cKSVD&Sparse Representation for Color Image Restoration&\citealt{MairalEtAl2008}\\
LSSC&Learned Simultaneous Sparse Coding&\citealt{MairalEtAl2009}\\
BKSVD&Bayesian K-SVD&\citealt{SerraEtAl2017}\\
\midrule
\multicolumn{3}{c}{\vspace{\spac}}\\
\midrule
\multicolumn{3}{|c|}{Other Probabilistic Methods (GMMs, MRFs)}\\
\midrule
EPLL&Expected Patch Log Likelihood&\citealt{ZoranWeiss2011}\\
PLE&Piecewise Linear Estimators&\citealt{YuEtAl2012}\\
FoE&Fields of Experts&\citealt{RothAndBlack2009}\\
MRF&Markov Random Fields&\citealt{SchmidtEtAl2010}\\
NLRMRF&Non-Local Range Markov Random Field&\citealt{SunAndTappen2011}\\
\midrule
\multicolumn{3}{c}{\vspace{\spac}}\\
\midrule
\multicolumn{3}{|c|}{Non-Local Image Processing Methods}\\
\midrule
BM3D&Block-Matching and 3D Filtering&\citealt{DabovEtAl2007}\\
NL&Patch-based Non-Local Image Interpolation&\citealt{Li2008}\\
WNNM&Weighted Nuclear Norm Minimization&\citealt{GuEtAl2014}\\
\midrule
\multicolumn{3}{c}{\vspace{\spac}}\\
\midrule
\multicolumn{3}{|c|}{Deep Neural Networks}\\
\midrule
MLP&Multi Layer Perceptron&\citealt{BurgerEtAl2012}\\
DnCNN-B&Denoising Convolutional Neural Network&\citealt{ZhangEtAl2017}\\
TNRD&Trainable Nonlinear Reaction Diffusion&\citealt{ChenAndPock2017}\\
MemNet&Deep Persistent Memory Network&\citealt{TaiEtAl2017}\\
IRCNN&Image Restoration Convolutional Neural Network&\citealt{ChaudhuryAndRoy2017}\\
FFDNet&Fast and Flexible Denoising Convolutional Neural Network&\citealt{ZhangEtAl2018}\\
DIP&Deep Image Prior&\citealt{UlyanovEtAl2018}\\
DPDNN&Denoising Prior Driven Deep Neural Network&\citealt{DongEtAl2019}\\
BDGAN&Image Blind Denoising Using Generative Adversarial Networks&\citealt{ZhuEtAl2019}\\
BRDNet&Batch-Renormalization Denoising Network&\citealt{TianEtAl2020}\\
\midrule
\end{tabular}
}
\caption{Algorithms used for comparsion on denoising and/or inpainting benchmarks. The algorithms
have been grouped into four types of approaches. The table includes the best performing
algorithms and standard baselines.}
\label{tab:other_ir_algos}
\end{table}
\begin{table}
\centering
\resizebox{\linewidth}{!}{
  \begin{tabular}{ ll }
  {\Large\bf A} Denoising & {\Large\bf B} Inpainting \\
  \begin{tabular}{|cc|c|c|c|}
  \hline
  \multicolumn{2}{|c|}{}&\multicolumn{3}{c|}{Further a-priori knowledge required}\\[5pt]
  &&\multirow{2}{*}{None}&\multirow{2}{*}{\makecell{Noise\\Level}}&\multirow{2}{*}{\makecell{Test-Train\\Match}}\\
  &&&&\\
  \cline{1-5}
  \multirow{18}{*}{\rotatebox[origin=c]{90}{Clean training data required}}&\multirow{7}{*}{\rotatebox[origin=c]{90}{No}}&\ReqLab{DE1}&\ReqLab{DE2}&\ReqLab{DE3}\\
  &&\multirow{7}{*}{\makecell{MTMKL\\BPFA\\GSC\\BKSVD\\EBSC\\ES3C}}&\multirow{7}{*}{\makecell{KSVD\\LSSC\\BM3D\\WNNM}}&\multirow{7}{*}{\makecell{\na}}\\
  &&&&\\
  &&&&\\
  &&&&\\
  &&&&\\
  &&&&\\
  &&&&\\
  \cline{2-5}
  &\multirow{9}{*}{\rotatebox[origin=c]{90}{Yes}}&\ReqLab{DE4}&\ReqLab{DE5}&\ReqLab{DE6}\\
  &&\multirow{9}{*}{\makecell{\na}}&\multirow{9}{*}{\makecell{EPLL}}&\multirow{9}{*}{\makecell{MLP\\DnCNN-B\\TNRD\\MemNet\\FFDNet\\DPDNN\\BDGAN\\BRDNet}}\\
  &&&&\\
  &&&&\\
  &&&&\\
  &&&&\\
  &&&&\\
  &&&&\\
  &&&&\\
  &&&&\\  
  \hline
  \end{tabular}&
  \begin{tabular}{|cc|c|c|c|}
  \hline
  \multicolumn{2}{|c|}{}&\multicolumn{3}{c|}{Further a-priori knowledge required}\\[5pt]
  &&\multirow{2}{*}{None}&\multirow{2}{*}{\makecell{Noise\\Level}}&\multirow{2}{*}{\makecell{Test-Train\\Match}}\\
  &&&&\\
  \cline{1-5}
  \multirow{18}{*}{\rotatebox[origin=c]{90}{Clean training data required}}&\multirow{7}{*}{\rotatebox[origin=c]{90}{No}}&\ReqLab{IN1}&\ReqLab{IN2}&\ReqLab{IN3}\\
  &&\multirow{7}{*}{\makecell{MTMKL\\BPFA\\BKSVD\\DIP\\ES3C}}&\multirow{7}{*}{\makecell{NL\\PLE}}&\multirow{7}{*}{\makecell{\na}}\\
  &&&&\\
  &&&&\\
  &&&&\\
  &&&&\\
  &&&&\\
  &&&&\\
  \cline{2-5}
  &\multirow{9}{*}{\rotatebox[origin=c]{90}{Yes}}&\ReqLab{IN4}&\ReqLab{IN5}&\ReqLab{IN6}\\
  &&\multirow{9}{*}{\makecell{FoE\\MRF}}&\multirow{9}{*}{\makecell{cKSVD}}&\multirow{9}{*}{\makecell{NLRMRF\\IRCNN}}\\
  &&&&\\
  &&&&\\
  &&&&\\
  &&&&\\
  &&&&\\
  &&&&\\
  &&&&\\
  &&&&\\
  \hline
  \end{tabular}\\
  \end{tabular}
}
  \caption{Algorithms for denoising and inpainting require different degrees and types of prior knowledge. 
In {\bf A} the algorithms listed in the column ``Noise Level'' require the ground-truth noise level of the test data as input parameter. The denoising approaches listed in the column ``Test-Train Match'' are optimized either for one single or for several particular noise levels. 
In {\bf B} the column ``Test-Train Match'' lists inpainting algorithms that are optimized for a particular percentage of missing pixels (e.g. for images with 80\% randomly missing values) or for a particular missing value pattern. These algorithms (category IN6) can be applied without providing a mask that indicates the pixels that are to be filled (see Appendix\,\ref{app:discussion_on_evaluation} for details).\vspace{-5mm}}
  \label{tab:requirements}
\end{table}

\subsection{Image Denoising}
\label{subsec:denoising}
To apply EBSC and ES3C for image denoising, we first preprocessed the noisy images that we aimed to restore by segmenting the images into smaller patches: given an image of size $\cal{H}\times \cal{W}$ and using a patch size of $D=P_x\times P_y$ we cut all possible $N=(\cal{W} - \cal{P}_{\text{x}} + \text{1})\times(\cal{H} - \cal{P}_\text{y} + \text{1})$ patches by moving a sliding window on the noisy image. Patches were collected individually for each image and corresponded to the datasets $\mathcal{Y}$ which EBSC or ES3C were trained on. In other words, EBSC and ES3C leveraged nothing but the information of the noisy image itself (no training on other images, no training on clean images, noise unknown a-priori).
After model parameters had been inferred using EBSC and ES3C, we took expectations \wrt the posterior predictive distribution of the model in order to estimate the non-noisy image pixels. For each noisy patch, we estimated all its non-noisy pixel values, and as commonly done we repeatedly estimated the same pixel value based on different (mutually overlapping) patches that shared the same pixel. The different estimates for the same pixel were then gathered to determine the non-noisy value using a weighted average (for details see Section\,\ref{subsec:estimator} and Appendix\,\ref{app:data_estimator}; also see, e.g., \citealt{BurgerEtAl2012}). Finally, we were interested in how well EBSC and ES3C could denoise a given image in terms of the standard peak-signal-to-noise ratio (PSNR) evaluation measure.

\subsubsection{Relation between PSNR measure and variational lower bound}
\label{subsec:denoising_psnr_free_energy}
\begin{figure}
\centering
\includegraphics{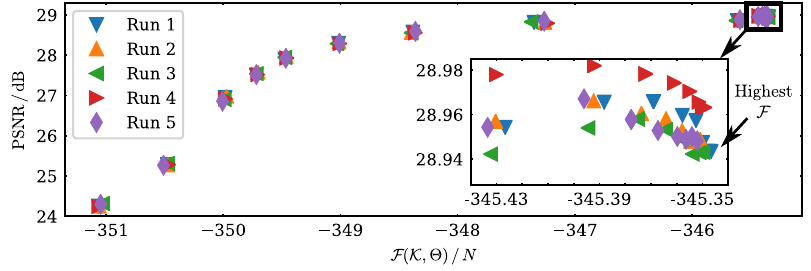}
\caption{Variational bound and PSNR value pairs obtained from applying ES3C to the noisy ``House'' image (AWG noise with $\sigma=50$). 2,000 EEM iterations were performed and PSNR values were measured at iterations 1, 2, 5, 10, 20, 50, 100, 200, 400, 600, 800, 1000, 1200, 1400, 1600, 1800, 2000. The experiment was repeated five times using the same noise realization in each run. The PSNR values at the last iteration were 28.94, 28.95, 28.94, 28.96, 28.95 (avg.\ rounded $28.95\pm0.01$; all values in~dB).}
\label{fig:ssscDenoising_house_psnr_fenergy}
\end{figure}

We first investigated whether the value of the learning objective (the variational lower bound) of EBSC and ES3C was instructive about the denoising performance of the algorithms in terms of PSNR. While the PSNR measure requires access to the clean target image, the learning objective can be evaluated without such ground-truth knowledge. If learning objective and PSNR measure were reasonably well correlated, one could execute the algorithms several times on a given image and determine the best run based on the highest value of the learning objective without requiring access to the clean image. 
For the experiment, we used the standard ``House'' image degraded by additive white Gaussian (AWG) noise with a standard deviation of $\sigma=50$ (the clean and noisy images are depicted in Figure\,\ref{fig:ssscDenoising_house_all_in_one}). We applied ES3C for denoising using patches of size $D=8\times8$ and a dictionary with $H=256$ elements (EEM hyperparameters listed in Table~\ref{tab:hyperparams} in Appendix\,\ref{app:numerical_experiments_tech}). At several iterations during execution of the algorithm, we measured the value of the learning objective and the PSNR of the reconstructed image. The experiment was repeated five times, each time using the same noisy image but a different initialization of the algorithm. Figure\,\ref{fig:ssscDenoising_house_psnr_fenergy} illustrates the result of this experiment.
As can be observed, PSNR values increase with increasing value of the objective $\mathcal{F}(\mathcal{K},\Theta)\,/\,N$ during learning. The truncated distributions
which give rise to the lower bound are consequently well suited. Only at the end of learning, PSNR values slightly decrease while the lower bound still increases (inset figure).
The final PSNR decreases are likely due to the PSNR measure being not perfectly correlated with the data likelihood. However, decreases of the likelihood while the lower bound still increases cannot be excluded. While final decreases of PSNR are small (see scale of inset figure of Figure\,\ref{fig:ssscDenoising_house_psnr_fenergy}), their consequence is that
the run with the highest final value for the variational bound is not necessarily the run with the highest PSNR. For Figure\,\ref{fig:ssscDenoising_house_psnr_fenergy}, for instance, Run~1 has the highest final variational bound but Run~4 the highest final PSNR.

\subsubsection{Comparison of Generative Models and Approximate Inference}
\label{subsec:denoising_controlled_conditions}
Before comparing EBSC and ES3C with a broad range of different denoising algorithms (further below), we first focused on approaches that are all based on essentially the same 
data model. Differences in measured performance can than be attributed more directly to differences in the method used for parameter optimization. 
Concretely, we here first consider algorithms that are all based on a spike-and-slab sparse coding (SSSC) model. Namely, we compare the following algorithms:\vspace{-1ex}
\begin{itemize}
  \item[-] the ES3C algorithm which is trained using variational optimization of truncated posteriors (EEM),\vspace{-1ex}
  \item[-] the MTMKL algorithm \citep{TitsiasLazaro2011} which uses a factored variation approach (i.e., mean field) for parameter optimization \citep[also compare][]{GoodfellowEtAl2012},\vspace{-1ex}
  \item[-] the BPFA algorithm \citep[][]{ZhouEtAl2012} which uses sampling for parameter optimization,\vspace{-1ex}
  \item[-] the GSC algorithm \citep[][]{SheikhEtAl2014} which uses truncated posteriors constructed using preselection (i.e., no variational loop).\vspace{-1ex}
\end{itemize}
Furthermore, we included EBSC into the comparison, as the BSC data model \citep[][]{HennigesEtAl2010} is a boundary case of the SSSC model. For best comparability, we investigated the denoising performance of the algorithms under controlled conditions: We considered a fixed task (the ``House'' benchmark with $\sigma=50$ AWG noise), and we compared algorithms for three different configurations of $D$ and $H$ (see Figure\,\ref{fig:ssscDenoising_house_all_in_one}\,B). PSNR values for BPFA, MTMKL and GSC are taken from the corresponding original publications. The publication for GSC only reports values for $D=8\times{}8$ and $H=64$ as well as for $D=8\times{}8$ and $H=256$; MTMKL only for the former and BPFA only for the latter setting. We additionally cite the performance of MTMKL for $D=8\times{}8$ and $H=256$ as reported by \citet[][]{SheikhEtAl2014}.
PSNR values for the algorithms EBSC and ES3C were obtained as described above (Section\,\ref{subsec:denoising}) using the same hyperparameters for both approaches (see Table~\ref{tab:hyperparams} in Appendix\,\ref{app:numerical_experiments_tech}). The PSNR values for EBSC and ES3C are averages across three runs each (standard deviations were smaller or equal 0.13\,dB PSNR).

Considering Figure\,\ref{fig:ssscDenoising_house_all_in_one}\,B, BPFA and MTMKL (which both represented the state-of-the-art at the time of their publication) perform well but in comparison ES3C shows significant further improvements.  
As the major difference between the considered approaches is the parameter optimization approach rather than the generative model, the performance increases of ES3C compared to the other algorithms can be attributed to the used EEM optimization. An important factor of the performance increases for ES3C is presumably its ability to leverage the available generative fields more effectively. \citet{TitsiasLazaro2011} report for MTMKL, for instance, that larger dictionaries ($H>64$) do not result in performance increases; similarly \citet{ZhouEtAl2012} report that BPFA models with 256 and 512 dictionary elements yield similar performance. GSC has been shown to make better use of larger dictionaries than MTMKL \citep[][]{SheikhEtAl2014}. Performance of GSC is, however, significantly lower than ES3C for all investigated dictionary sizes, which provides strong evidence for the advantage of the fully variational optimization used by EEM. The EEM optimization also results in a notably strong performance of EBSC compared to previous approaches: it outperforms MTMKL (as well as GSC for $H=64$). The strong performance may be unexpected as the underlying BSC model is less expressive than the SSSC data model. The performance gains using EEM thus outweigh the potentially better data model in this case. However, ES3C performs (given the same set of hyperparameters) much stronger than EBSC in all settings, and can make much more use of larger dictionaries and larger patches.

\subsubsection{General comparison of denoising approaches}
\label{subsec:denoising_general_benchmarking}
After having compared a selected set of algorithms under controlled conditions, we investigated denoising performance much more generally. We compared the best performing algorithms on standard denoising benchmarks irrespective of the general approach they follow, allowing for different task categories they require, and allowing for any type of hyperparameters they use. The methods we have used for comparison are listed in Table\,\ref{tab:requirements}\,A, where the algorithms are grouped according to the prior information they assume (compare Section\,\ref{subsec:other_ir_algos} and Appendix\,\ref{app:discussion_on_evaluation}). 

\begin{figure}
\centering
\vspace{-0.3cm}
\resizebox{\linewidth}{!}{
\includegraphics{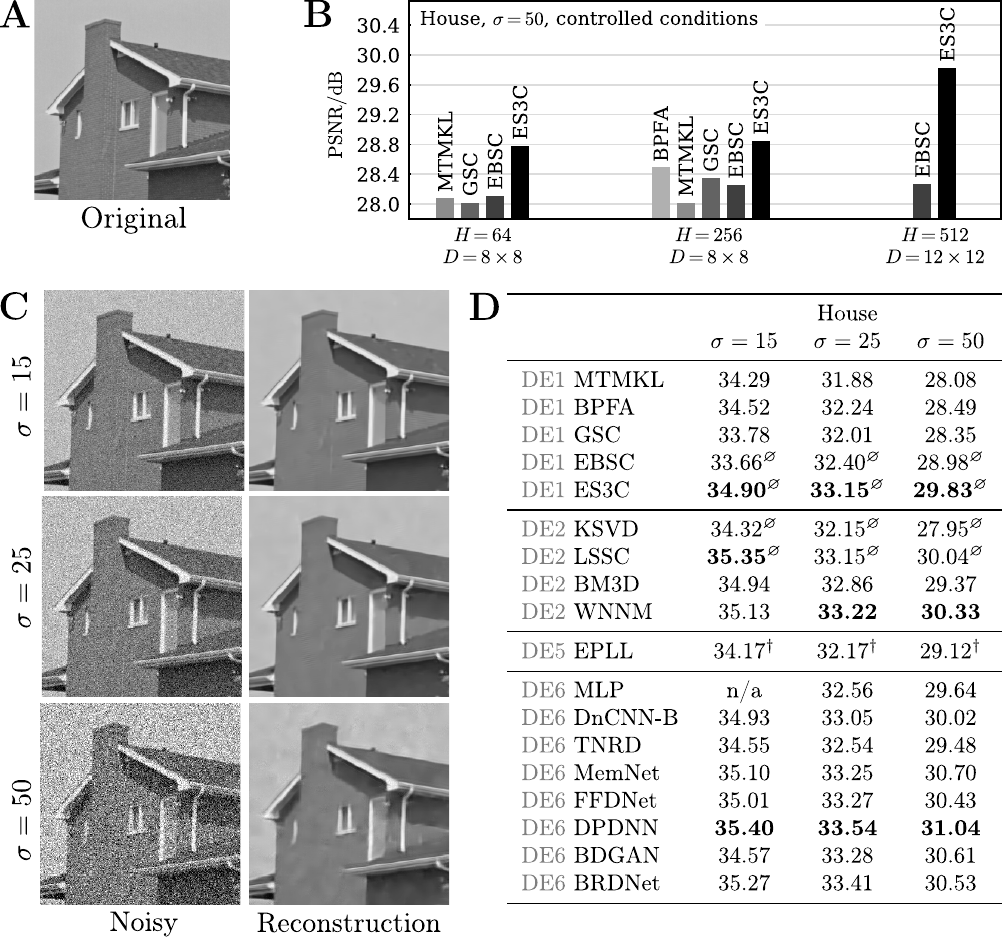}
}
\caption{Denoising results for the ``House'' image (clean original depicted in {\bf A}). {\bf B} Performance comparison for AWG noise with $\sigma=50$ under controlled conditions (see text for further details). {\bf D} Performance comparison \wrt to state-of-the-art denoising approaches and standard baselines using different optimized hyperparameters (see text for further details). Grouping and labeling according to Table\,\ref{tab:requirements}; in each group the highest PSNR value is marked bold. Numbers marked with \AvgPSNR correspond to averages over multiple independent realizations of the experiment using different realizations of the noise (see text for further details). For EPLL, model selection was performed based on the learning objective of the algorithm (\BestPSNRwrtObj markers; personal communication with D. Zoran). Panel {\bf C} illustrates the reconstructed images obtained with ES3C in the run with the highest PSNR value.}
\label{fig:ssscDenoising_house_all_in_one}
\end{figure}

A limitation of most of the investigated algorithms is the computational resources they require. For EBSC and ES3C, model size (i.e., patch and dictionary sizes) are limited by available computational resources and so is the number of variational states $S$ (Section\,\ref{subsec:evolutionary_optimization}) used to approximate posteriors. For ES3C, a good trade-off between performance vs.\ computational demand was observed for instance for $D=12\times12$ and $H=512$ with $S=60$ for the experiment of Figure\,\ref{fig:ssscDenoising_house_all_in_one}\,D; 
for EBSC a good trade-off between performance and computational demand was observed for $D=8\times8$ and $H=256$ with $S=200$. We thus have used more variational states $S$ for EBSC than for ES3C. Technical details including EEM hyperparameters are provided in Appendix\,\ref{app:numerical_experiments_tech}.
\begin{figure}
  \centering
  \resizebox{\linewidth}{!}{
  \includegraphics{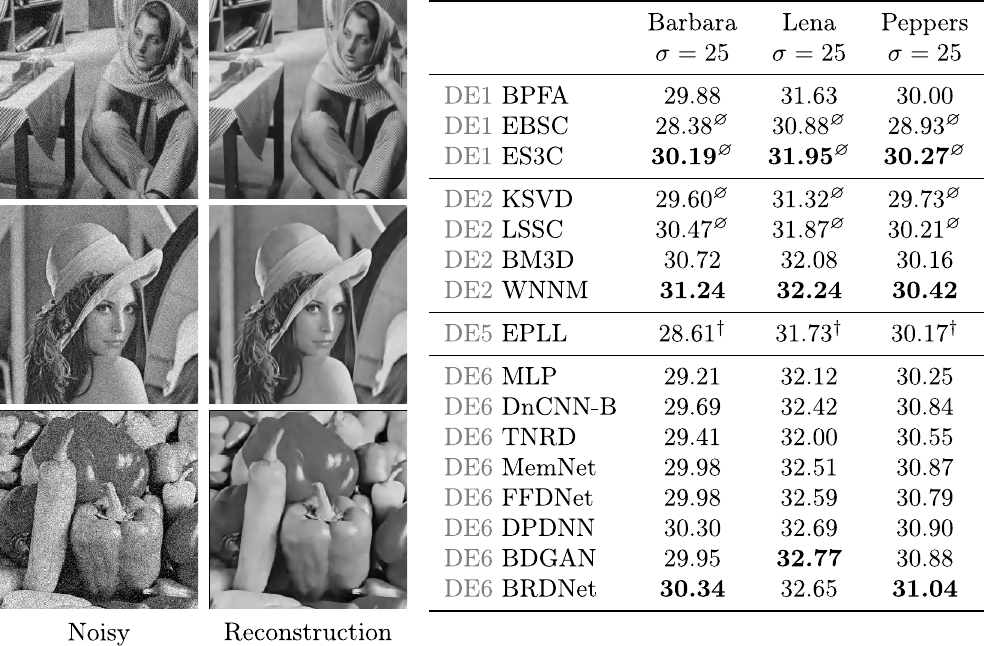}
  }
  \caption{Denoising results for ``Barbara'', ``Lena'' and ``Peppers''. The table shows a performance comparison \wrt to state-of-the-art denoising approaches and standard baselines using different optimized hyperparameters (see text for further details). Grouping and labeling according to Table\,\ref{tab:requirements}; in each group the highest PSNR value is marked bold. Numbers marked with \AvgPSNR correspond to averages over multiple independent realizations of the experiment using different realizations of the noise (see text for further details). For EPLL, model selection was performed based on the learning objective of the algorithm (\BestPSNRwrtObj markers; personal communication with D. Zoran). On the left we illustrate the reconstructed images obtained with ES3C in the run with the highest PSNR value.}
  \label{fig:ssscDenoising_barblenpep}
\end{figure}

Figures\,\ref{fig:ssscDenoising_house_all_in_one}\,D and \ref{fig:ssscDenoising_barblenpep} list denoising performances obtained with algorithms from the different categories of Table\,\ref{tab:requirements}. 
The listed PSNR values are taken from the respective original publications with the following exceptions: For WNNM, EPLL and MLP, values are taken from \citet{ZhangEtAl2017} and for TNRD and MemNet values equal the ones in \citet{DongEtAl2019}. 
PSNR values reported in the original papers are often single PSNR values which may correspond to a single run of the investigated algorithm or to the best observed PSNR value. For deterministic methods with essentially no PSNR variations for a fixed input, it might seem natural to report a single PSNR value. For stochastic methods, reporting the best observed value is instructive about how well a given method can in principle perform. In practice, however, the run with
highest PSNR value is less relevant than the average PSNR as selecting the run with best PSNR usually requires ground-truth (compare discussion in Appendix\,\ref{app:discussion_on_evaluation}).
For KSVD and LSSC as well as for EBSC and ES3C, the PSNR values listed in Figures\,\ref{fig:ssscDenoising_house_all_in_one}\,D and \ref{fig:ssscDenoising_barblenpep} correspond to averages. For the two former approaches, averages were obtained from five runs of the algorithms with different noise realizations (standard deviations were reported to be negligibly small; see \citealt{EladAndAharon2006}, \citealt{MairalEtAl2009}). For EBSC and ES3C, we performed three runs of the algorithms with different noise realizations (observed standard deviations of the resulting PSNRs were not larger than 0.06\,dB; compare Table\,\ref{tab:app_den_inp_all_runs} in Appendix\,\ref{app:numerical_results}).

Considering Figures\,\ref{fig:ssscDenoising_house_all_in_one}\,D and \ref{fig:ssscDenoising_barblenpep}, it can be observed that the performance tends to increase the more a-priori information is used by an algorithm (as can be expected). Feed-forward DNNs report overall the best PSNR values but also use the most a-priori information (see category DE6, Table\,\ref{tab:requirements}). In the ``zero-shot'' category with no a-priori information (DE1), ES3C shows the best performance (Figure\,\ref{fig:ssscDenoising_house_all_in_one}\,D). ES3C significantly increases the state-of-the-art PSNR values across all investigated settings.
ES3C is still competitive if compared to algorithms in categories DE2 and DE5 which use more a-priori information but it is outperformed by some of these methods especially at lower noise levels (Figures\,\ref{fig:ssscDenoising_house_all_in_one}\,D and \ref{fig:ssscDenoising_barblenpep}). 
Notably, differences in performance between the categories get smaller for larger and more complex images (Figure\,\ref{fig:ssscDenoising_barblenpep}). For the image ``Barbara'', for instance, ES3C outperforms all DNNs except of DPDNN and BRDNet which are slightly better while the best result is obtained for the image specific non-local method WNNM.

PSNR values achieved by the previous methods of BPFA or MTMKL have notably proven to be difficult to further improve. Only systems using more a-priori information (such as WNNM in category DE2 or DNNs in category DE6) have more recently reported higher PSNR values \citep[also see discussion of difficult to improve BM3D values by][]{GuEtAl2014}. In category DE1, the only more recent competitive approach is a Bayesian version of K-SVD \citep[BKSVD;][]{SerraEtAl2017}. By extending K-SVD using Bayesian methodology, the data noise can be estimated from data. The BKSVD approach can thus be applied without a-priori information
on the noise level, which makes it applicable in the ``zero-shot'' category \citep[see][for a discussion]{SerraEtAl2017}. PSNR values for BKSVD reported by \citealt[][]{SerraEtAl2017} are computed on downscaled images, and are therefore not directly comparable to those of Figures\,\ref{fig:ssscDenoising_house_all_in_one}\,D and \ref{fig:ssscDenoising_barblenpep}. However, comparisons of BKSVD with MTMKL, K-SVD and BPFA reported by \citet[][]{SerraEtAl2017} show that BKSVD performs competitive especially for lower noise levels. For higher noise levels, though, (e.g.\ $\sigma=25$ for `Lena' and `Peppers') BPFA results in higher PSNR values (and in similar PSNRs for `Barbara'). As ES3C outperforms BPFA for $\sigma=25$ on `Barbara', `Lena' and `Peppers' on the full resolution images, BKSVD also does not seem to establish a new state-of-the-art in
the DE1 category. We can, therefore, conclude that ES3C establishes a novel state-of-the-art in the ``zero-shot'' category since a relatively long time.


\subsection{Image Inpainting}
\label{subsec:inpainting}
The final benchmark we used is image inpainting, i.e., the reconstruction of missing values in images. 
A standard setting for this task is the restoration of images that have been artificially degraded by uniformly randomly deleting pixels. We here considered the standard test images ``Barbara'', ``Cameraman'', ``Lena'',  ``House'' and ``Castle'' as examples for gray-scale and RGB images.
For denoising, ES3C consistently performed better than EBSC, and the same we observed for inpainting. The performance difference between ES3C and EBSC on inpainting was, however, 
larger (sometimes several dB). We therefore focus on ES3C for comparison as EBSC is not competitive to the best performing methods we compare to.

For inpainting with ES3C, we employed the same partitioning and averaging scheme as described in Section\,\ref{subsec:denoising}. For the RGB image, ES3C was applied jointly to all color channels, as this had been reported to be more beneficial than training individual models on different color layers \citep{MairalEtAl2008}. We trained the model and restored missing values exclusively based on the incomplete data which we aimed to restore (following, e.g., \citealt{FadiliStarck2005,LittleRubin1987}); uncorrupted pixels were not modified. Details about the missing value estimator applied by ES3C are provided in Section\,\ref{subsec:estimator} and in Appendix\,\ref{app:data_estimator}. For inpainting, we used a SSSC model with fixed $\muVec=\vec{1}$ and $\Psi=\eye$ as these parameters were observed to grow exponentially during learning which led to numerical instabilities of the training process. 
For each test scenario, we performed three runs of ES3C and used a different realization of missing values in each run (EEM hyperparameters are listed in Table\,\ref{tab:hyperparams} in Appendix\,\ref{app:numerical_experiments_tech}). Compared to the denoising experiments (Section\,\ref{subsec:denoising_general_benchmarking}), we observed for inpainting slightly larger variations of the resulting PSNR values (standard deviations were smaller or equal 0.18\,dB; results of individual runs listed in Table\,\ref{tab:app_den_inp_all_runs} in Appendix\,\ref{app:numerical_results}).
\begin{figure}
\centering
\resizebox{\linewidth}{!}{
\includegraphics{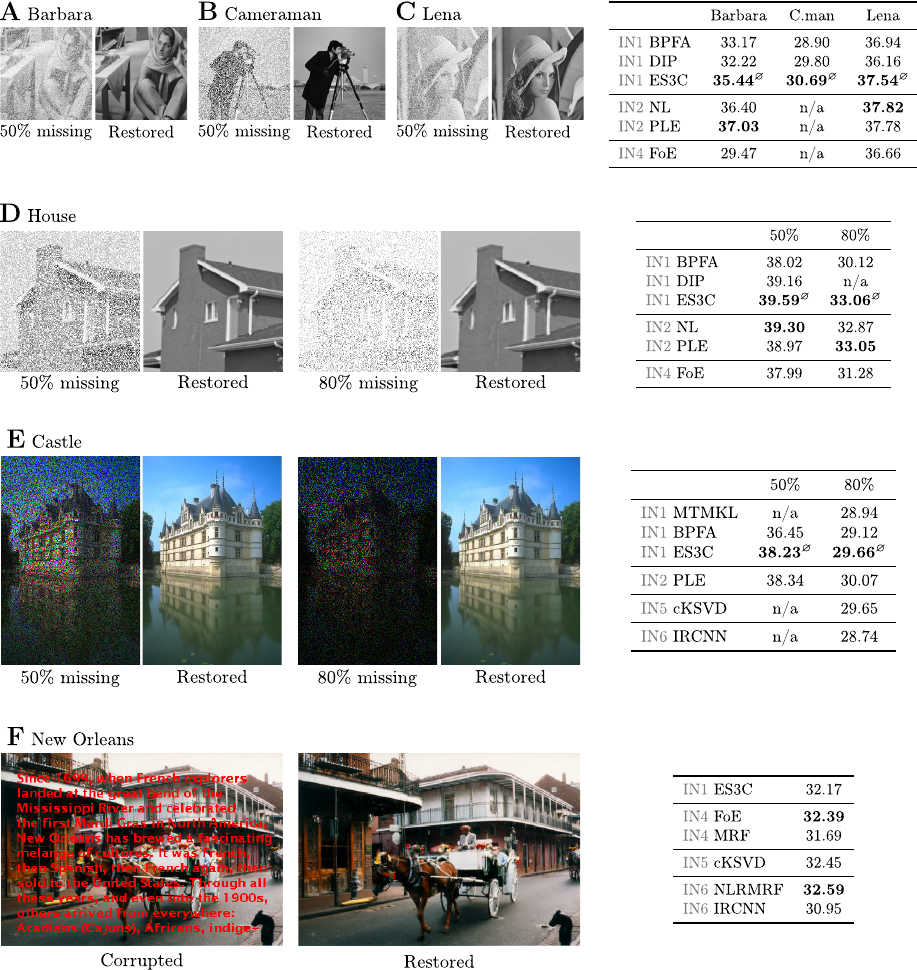}
}
\caption{Inpainting results for ``Barbara'', ``Cameraman'', ``Lena'', ``House'', ``Castle'' and ``New Orleans''. For the performance comparison, different approaches are grouped and labeled according to Table\,\ref{tab:requirements}. In each group, the highest PSNR value is marked bold. 
Numbers marked with \AvgPSNR in {\bf A} to {\bf E} correspond to averages over multiple independent runs of the experiment using different realizations of missing values (see text for details; the performance of ES3C in the individual runs is reported in Table\,\ref{tab:app_den_inp_all_runs} in Appendix\,\ref{app:numerical_results}). On the left we illustrate the reconstructed images obtained with ES3C in the run with the highest PSNR value.}
\label{fig:ssscInpainting}
\end{figure}

The inpainting results for ``Barbara'', ``Cameraman'', ``Lena'' (each 50\% missing values) and ``House'' and ``Castle'' (each 50\% and 80\% missing values) are depicted in Figure\,\ref{fig:ssscInpainting}\,A-E. The reference PSNR values are taken from the original publications except for NL and FoE for which we cite the numbers as reported by \citet{YuEtAl2012}. In general, inpainting is a less wide-spread benchmark than denoising presumably because additional estimation routines have to be developed and missing values pose, more generally, a considerable challenge to many approaches. 
Considering Figure\,\ref{fig:ssscInpainting}\,A-E, first observe that performance differences between categories are larger than for denoising. Compared to the best performing approaches in the literature, ES3C shows to be competitive in many settings. On the ``Castle'' benchmark, ES3C performs better than all other approaches in the literature except of PLE which uses the noise level as a-priori information. For the ``House'' benchmark, ES3C establishes a novel state-of-the-art in general: it performs better than all other algorithms for $50\%$ lost pixels and equal to PLE for $80\%$ lost pixels (i.e., within PSNR standard deviation of ES3C).
This novel state-of-the-art is notably reached without requiring clean training data or a-priori knowledge about noise level or sparsity (ES3C
requires to know which pixels are missing; compare Table\,\ref{tab:requirements} and Appendix\,\ref{app:discussion_on_evaluation}).

Finally, we also report inpainting performance on the well-known ``New Orleans'' inpainting benchmark (Figure\,\ref{fig:ssscInpainting}\,F left-hand-side). The benchmark serves as
an inpainting example where pixels are lost non-randomly. For the benchmark, the original image is artificially corrupted by adding overlaid text, which is then removed by applying inpainting. Compared to inpainting with randomly missing values (e.g., the benchmarks in Figure\,\ref{fig:ssscInpainting}\,A-E), the ``New Orleans'' image contains relatively large contiguous regions of missing values. 
For our measurements, we used a publicly available dataset\footnote{We downloaded the clean and the corrupted image from https://lear.inrialpes.fr/people/mairal/resources/ KSVD\_package.tar.gz and the text mask from ttps://www.visinf.tu-darmstadt.de/media/visinf/ software /foe\_demo-1\_0.zip; all images were provided as PNG files. We verified that the corrupted image was consistent \wrt the mask and the original image contained in the dataset: We applied the mask to the original image and compared the unmasked parts to the unmasked parts of the corrupted image. These were identical. After filling masked areas with red (i.e., replacing the corresponding pixels with (255,0,0) in RGB space) we measured a PSNR value of the corrupted image of 13.48\,dB. This value slightly deviates from numbers reported in other studies; \citeauthor{ChaudhuryAndRoy2017} for example measured a PSNR of 15.05~dB.}. 
The result of the experiment is illustrated in Figure~\ref{fig:ssscInpainting}\,F. Reference PSNR values are taken from respective original publications with exceptions of FoE and MRF for which we use values reported by \cite{SunAndTappen2011}. Due to extensive runtimes, we report the result of a single run of ES3C for this benchmark (see Appendix\,\ref{app:numerical_experiments_tech} for further details; EEM hyperparameters listed in Table~\ref{tab:hyperparams}). 
As can be observed, ES3C also performs well on this task. We observed higher PSNR values for ES3C than for MRF and IRCNN. Values for FoE and cKSVD are higher but also more a-priori information is used. Especially FoE is less sensitive to larger areas of lost pixels than ES3C (compare the lower performance of FoE in Figure\,\ref{fig:ssscInpainting}\,A-D); presumably FoE can make better use of larger contexts. 
%
%
The state-of-the-art on this benchmark is established by NLRMRF.

\section{Discussion}
Efficient approximate inference is of crucial importance for the training of expressive generative models. Consequently, there exist a range of different methods with different assumptions and different features. As most generative models can only be trained using approximations and usually many locally optimal solutions exist, the question of how well a generative model can potentially perform on a given task is difficult to answer.
Sophisticated, mathematically grounded approaches such as sampling or variational optimization have been developed in order to derive sufficiently precise and efficient learning algorithms. As a contribution of this work, we have proposed a general purpose variational optimization that directly and intimately combines evolutionary algorithms (EAs) and EM. Considering Algorithm\,\ref{alg:eem}, EAs are an integral part of variational EM where they address the key optimization problem (the variational loop) arising in the training of directed generative models. The EEM algorithm is defined by a set of optimization hyperparameters for the EA. Given the hyperparameters, EEM is applicable to a given generative model solely by using the model's joint probability $p(\sVec,\yVec\,|\,\Theta)$. No analytical derivations are required to realize variational optimization.\vspace{1mm}
\\
\\
\noindent{\bf Relation to other approximate inference approaches.} To show large-scale applicability of EEM, we considered high-dimensional image data and generative models with large (combinatorial) state spaces. Aside from the elementary generative models of noisy-OR Bayes Nets \citep[][]{SingliarHauskrecht2006,JerniteEtAl2013,RotmenschEtAl2017} and binary sparse coding \citep[][]{HaftEtAl2004,HennigesEtAl2010}, we used spike-and-slab sparse coding (SSSC) as a more expressive example.
The SSSC data model has been of considerable interest due to its strong performance for a range of tasks \citep[][]{ZhouEtAl2009,ZhouEtAl2012,TitsiasLazaro2011,GoodfellowEtAl2012,SheikhEtAl2014}. However, its properties require sophisticated probabilistic approximation methods for parameter optimization. Essentially three types of approximations have previously been applied to train the model: mean field \citep{TitsiasLazaro2011,GoodfellowEtAl2012}, truncated posterior approximations \citep{SheikhEtAl2014}, and MCMC sampling \citep{ZhouEtAl2009,ZhouEtAl2012,MohamedEtAl2012}. The EEM method
shares features with all three previous approaches: like work by \citet{SheikhEtAl2014}, EEM uses truncated posterior distributions to approximate full posteriors, which contrasts with factored posterior approximations (i.e., mean field) used by \citet[][]{TitsiasLazaro2011,GoodfellowEtAl2012}.
However, like work by \citeauthor{TitsiasLazaro2011}, \citeyear{TitsiasLazaro2011}, and \citeauthor{GoodfellowEtAl2012}, \citeyear{GoodfellowEtAl2012}, EEM is a fully variational EM algorithm which is guaranteed to monotonically increase the variational lower bound. This contrasts with \citet{SheikhEtAl2014} or \citet{SheltonEtAl2017} who used feed-forward estimates to construct approximate posteriors (without a guarantee to monotonically increase a variational bound). Finally, like sampling based approaches \citep{ZhouEtAl2009,ZhouEtAl2012,MohamedEtAl2012}, EEM computes estimates of posterior expectations based on a finite set of latent states. However, unlike for sampling approximations, these states are not samples but are themselves variational parameters that are evolved using EAs (with a fitness defined by a variational bound).
\\
\\
\noindent{\bf Denoising and inpainting benchmarks and comparison with other approaches.} For comparison with a range of other methods, we used standard denoising and inpainting benchmarks and evaluated EEM for the SSSC model (termed {\em ES3C}) and EEM for the BSC model (termed {\em EBSC}). 
%
%
Compared to competing methods in the ``zero-shot'' category (see category DE1 and IN1 in Table\,\ref{tab:requirements}), we observed ES3C to achieve the highest PSNR values on all of the benchmarks we performed. 
Furthermore, ES3C also shows improvements w.r.t\ methods that use more a-priori information in many test scenarios:
\begin{itemize}
\item[-] The denoising results (Figures\,\ref{fig:ssscDenoising_house_all_in_one}\,-\,\ref{fig:ssscDenoising_barblenpep}) show that ES3C can outperform state-of-the-art sparse coding and non-local approaches including, e.g., LSSC and BM3D which require ground-truth noise level information. ES3C also improves on the popular GMM-based EPLL method, which additionally requires clean training data. Furthermore, ES3C shows improvements compared to a number of recent, intricate DNNs which are tailored to image denoising and require (large amounts of) clean data for training. 
For instance, on the ``Barbara'' image (Figure\,\ref{fig:ssscDenoising_barblenpep}), ES3C is only outperformed by the very recent networks DPDNN \citep{DongEtAl2019} and BRDNet \citep{TianEtAl2020}.
At the same time, we note that the image noise for the benchmarks is Gaussian. Gaussian noise is commonly encountered for real data but it also makes the SSSC model well suited for this task in general.
\item[-] For image inpainting, we observed ES3C to provide state-of-the-art results also in general (i.e. across all categories) in some settings: compared to the best performing algorithms on the ``House'' benchmark for inpainting, ES3C achieves the highest PSNR value (Figure\,\ref{fig:ssscInpainting}\,D, best value shared with PLE for 80\% lost pixels). The inpainting benchmark is, however, much less frequently applied than its denoising counterpart for the same image because a treatment of missing data is required.
\end{itemize}

\noindent{}The observation that probabilistic approaches such as SSSC-based algorithms perform well on the inpainting benchmarks of Figure\,\ref{fig:ssscInpainting}\,A-E may not come as too much of a surprise.
Because for randomly missing values, the information provided by relatively small patches contains valuable information which an SSSC model can leverage well. 
Larger areas of missing values like studied using the ``New Orleans'' benchmark require larger contexts, whose information may be better leveraged by less local methods such as Markov Random Fields (Figure\,\ref{fig:ssscInpainting}\,F).
%

To answer the question of why EEM results in improvements, direct comparison with other optimization methods for the same or similar models are most instructive (Figure\,\ref{fig:ssscDenoising_house_all_in_one}\,B).
In general, the type of variational optimization can have a strong influence on the type of generative fields that are learned, e.g., for the SSSC model. Mean field methods (i.e., fully factored variational distributions) have a tendency to bias generative fields to be orthogonal \citep[][]{MacKay2001,IlinValpola2005}.
Such effects are also discussed for other generative models \citep[e.g.][]{TurnerSahani2011a,VertesSahani2018}. Biases introduced by a variational approximation can thus lead to increasingly suboptimal representations (see, e.g., \citeauthor{TitsiasLazaro2011}, \citeyear{TitsiasLazaro2011}, \citeauthor{SheikhEtAl2014}, \citeyear{SheikhEtAl2014}, for discussions).
Sampling is in general more flexible but practical implementations may (e.g., due to insufficient mixing) also bias parameter learning towards learning posteriors with single modes. The denoising benchmark under controlled conditions shows that the sampling based BPFA approach \citep[][]{ZhouEtAl2012} performs, for instance, better than the mean field approach MTMKL \citep[][]{TitsiasLazaro2011} and the truncated EM approach GSC \citep[][]{SheikhEtAl2014} but worse than ES3C which is based on EEM.
\\
\\
\noindent{\bf Comparison with other generative approaches.} Observe that for our denoising benchmarking in Figures\,\ref{fig:ssscDenoising_house_all_in_one} and \ref{fig:ssscDenoising_barblenpep}, the recent BDGAN approach is the only deep generative model. The reason is that except for BDGAN, neither for generative adversarial nets \citep[GANs;][]{GoodfellowEtAl2014} nor for variational autoencoders \citep[VAEs;][]{RezendeEtAl2014,KingmaWelling2014} competitive denoising or inpainting results are reported on these standard benchmarks (for GANs and VAEs, benchmarks other than those in Figures\,\ref{fig:ssscDenoising_house_all_in_one} to \ref{fig:ssscInpainting} are often considered; see Appendix\,\ref{app:comparison_to_gans_vaes} for further discussion).
%
%
The BDGAN approach \citep[][]{ZhuEtAl2019} does provide benchmark values for the denoising task. It shows competitive performance but like feed-forward DNNs requires large image corpora for training. At least in principle, GANs and VAEs are also applicable to the ``zero-shot'' category (DE1 and IN1 in Figures\,\ref{fig:ssscDenoising_house_all_in_one}\,-\,\ref{fig:ssscInpainting}), however. The BDGAN approach uses large and intricate DNNs whose parameters are presumably difficult to train using just one image, which may explain why it was not applied in the ``zero-shot'' category. Recent work by, e.g.\ \citet[][]{ShocherEtAl2018}, explicitly discusses the DNNs sizes and the use of small DNNs for ``zero-shot'' super-resolution.  

For VAEs, neither data for the ``zero-shot'' category nor for the other categories are (to the knowlege of the authors) available for the standard benchmarks we used. A reason may be that (like for the BDGAN) very intricate DNNs as well as sophisticated sampling and training methods are required to be competitive. Experiments with standard VAE setups that we conducted did, e.g.\ for denoising, not result in PSNR values close to those reported in Figures\,\ref{fig:ssscDenoising_house_all_in_one} and \ref{fig:ssscDenoising_barblenpep}. Another possible reason for the absence of competitive values for VAEs may, however, be related to the variational approximation used for VAEs.
VAEs for continuous data usually use Gaussian as variational distributions \citep[][]{RezendeEtAl2014,KingmaWelling2014} that are in addition fully factored (i.e., mean field). The parameters of the VAE decoder may thus suffer from similar biasing effects as they were described for mean field approaches as used, e.g., by MTMKL \citep[][]{TitsiasLazaro2011} for the SSSC model. That standard VAEs {\em do} show such biases has recently been pointed out, e.g., by \citet[][]{VertesSahani2018}. 

Like other generative models, VAEs (which use DNNs as part of their decoder and encoder) are, consequently, likely to strongly profit from a more flexible encoding of variational distributions. In the context of deep learning, such flexible variational distributions have been suggested, e.g., in the form of normalizing flows \citep[][]{RezendeMohamed2015}. In its standard version \citep[][]{RezendeMohamed2015} normalizing flows are making training significantly more expensive, however. 
Approximate learning using normalizing flows is centered around a sampling-based approximating of expectation values \wrt posteriors \citep[][]{RezendeMohamed2015}; but also compare related approaches \citep[][]{HuangEtAl2018,BigdeliEtAl2020}. By following such a strategy, samples of increasingly complex variational distributions can be generated by successively transforming samples of an elementary distribution.
The idea of successive transformations has also been used to directly parameterize the data distribution in what is now termed flow-based models \citep[e.g.][]{DinhEtAl2017,KingmaDhariwal2018}. An early example \citep[][]{DinhEtAl2014} is a form of non-linear ICA \citep[][]{HyvarinenPajunen1999}.
%
By using EAs, the EEM approach does, in contrast, follow a strategy that is notably very different from normalizing flows and other approaches. A consequence of these different strategies are very different sets of hyperparameters: EEM hyperparameters are centered around parameters for the EA such as population size as well as mutation, crossover and selection operators; hyperparameters of normalizing flows and related methods are centered around parameterized successive mappings (e.g., type of mapping, number transformations) and (if applicable) hyperparameters for sampling. Furthermore, and most importantly for practical applications, the types of generative models addressed by EEM and normalizing flow are essentially complementary: EEM focuses on generative models with discrete latents while normalizing flows and others focus on models with continuous latents.  \vspace{1mm}
\\
\\
\noindent{\bf Conclusion.} More generally, EEM can be regarded as an example of the significant potential in improving the accuracy of posterior approximations. Many current research efforts are focused on making parameterized models increasingly complex: usually an intricate DNN is used either directly (e.g., feed-forward DNNs/CNNs for denoising) or as part of a generative model. At least for tasks such as denoising or inpainting, our results suggest that improving the flexibility of approximation methods may be as effective to increase performance as improving model complexity. While EEM shares the goal of a flexible and efficient posterior approximation with many other powerful and successful methods, it has a distinguishing feature: a direct link from variational optimization to a whole other research field focused on optimization: evolutionary algorithms. For generative models with discrete latents, new results in the field of EAs can consequently be leveraged to improve the training of future (elementary and complex) generative models.

%


\paragraph{Acknowledgments}\mbox{}\\[2mm]
\noindent
We acknowledge funding by the DFG project HAPPAA (SFB 1330, B2), the BMBF project SPAplus (05M20MOA), the BMBF Gentner Scholarship awarded to EG, a grant for computing resources by the HLRN HPC-network (project nim00006) and support by the CARL HPC-cluster of Oldenburg University which is funded by the DFG through its Major Research Instrumentation Programme (INST 184/157-1 FUGG) and the Ministry of Science and Culture (MWK) of the Lower Saxony State.

\begin{appendices}
\appendix
\section{Log-pseudo joint formulation}
\label{app:lpj}
For efficient computation of log-joint probabilities $\log \genmodel$ (which are required to compute truncated expectations \eqref{eq:tvem_truncated_expectations} and the variational lower bound \eqref{eq:tvem_free_energy}), we introduce the following reformulation. When computing $\log \genmodel$ for a concrete model, we identify the terms that only depend on the model parameters $\Theta$ (and not on the data points $\yVec$ and the variational states $\sVec$). These terms, denoted $\mathrm{C}(\Theta)$ in the following, only need to be evaluated once when computing $\log \genmodel$ for different data points and for different variational states. We refer to the remaining terms that depend on both $\sVec$, $\yVec$ and $\Theta$ as {\it log-pseudo joint} and denote this quantity by $\widetilde{\log p}(\sVec,\yVec\mid\Theta)$. The log-joint probability $\log \genmodel$ can then be reformulated as:
\begin{align}
\log \genmodel = \widetilde{\log p}(\sVec,\yVec\mid\Theta) + \mathrm{C}(\Theta)
\end{align}
The concrete expressions of $\mathrm{C}(\Theta)$ and $\widetilde{\log p}(\sVec,\yVec\mid\Theta)$ for the models introduced in Section\,\ref{subsec:models} are listed below:
\\
\\
{\it Noisy-OR.}
\begin{align}
\mathrm{C}(\Theta) &= \sum_{h=1}^{H} \log(1-\pi_h)\\
\widetilde{\log p}(\sVec,\yVec\mid\Theta)
  &= \sum_{d=1}^{D} \Big( y_d \log \big(N_d(\sVec)\big) + (1 - y_d)\log\big(1-N_d(\sVec)\big) \Big)
  + \sum_{h=1}^{H} s_h \log\Big(\frac{\pi_h}{1-\pi_h}\Big)\\
   N_d(\sVec\,) &= 1 - \prod_{h=1}^{H}(1 - W_{dh}s_h)
\end{align}
with the special exception of $\sVec = \vec{0}$ for which
\begin{align}
\widetilde{\log p}(\sVec = \vec{0}, \yVec) =
\begin{cases}
   0 &\text{if}\,\, \yVec = \vec{0}\\
   -\inf &\text{otherwise}
\end{cases}
\end{align}
For practical computations, we set $\widetilde{\log p}$ to an arbitrarily low value rather than the floating point
infinite representation.
\\
\\
{\it BSC.} 
\begin{align}
\mathrm{C}(\Theta) &= H\log(1-\pi) - \frac{D}{2}\log(2\pi\sigma^2)\\
\widetilde{\log p}(\sVec,\yVec\mid\Theta) &= \log\Big(\frac{\pi}{1-\pi}\Big)\sum_{h=1}^{H} s_h - \frac{1}{2\sigma^2}(\yVec-W\sVec\,)\transp(\yVec-W\sVec\,)
\end{align}
\\
{\it SSSC.} 
\begin{align}
\mathrm{C}(\Theta) &= \sum_{h=1}^{H} \log(1-\pi_h) - \frac{D}{2}\log(2\pi)\\
\widetilde{\log p}(\sVec,\yVec\mid\Theta) &= \sum_{h=1}^{H} s_h\log\Big(\frac{\pi_h}{1-\pi_h}\Big) - \frac{1}{2}\log\vert C_{\sVec}\vert-\frac{1}{2}(\yVec-\tilde{W}_{\sVec\,}\muVec)\transp C_{\sVec}^{-1}(\yVec-\tilde{W}_{\sVec\,}\muVec)
\end{align}

\section{Sparsity-driven bitflips}
\label{app:eem_sparseflips}
When performing sparsity-driven bitflips, we flip each bit of a particular child $\sVec$ with probability $p_0$ if it is 0, with probability $p_1$ otherwise. We call $p_{\mathrm{bf}}$ the average probability of flipping any bit in $\sVec$.
We impose as constraints on $p_0$ and $p_1$ that $p_1 = \alpha p_0$ for some constant $\alpha$ and that the average number of ``on'' bits after mutation is set to $\widetilde{s}$. This yields the following expressions for $p_0$ and $p_1$:
\begin{gather}
p_0 = \frac{H p_{\mathrm{bf}}}{H + (\alpha-1) |\sVec\,|}, \ \ \ \ \ \ \ \ \ \ \ \ \ \  \alpha = \frac{(H - |\sVec\,|) \cdot (H  p_{\mathrm{bf}}  - \widetilde{s} + |\sVec\,|) }{( \widetilde{s} - |\sVec\,| + H  p_{\mathrm{bf}} ) |\sVec\,|}
\end{gather}
Trivially, random uniform bitflips correspond to the case $p_0 = p_1 = p_{\mathrm{bf}}$. 
For our numerical experiments (compare Section\,\ref{subsec:verification}), we chose $\widetilde{s}$ based on the sparsity learned by the model (we set $\widetilde{s}=\sum_{h=1}^{H} \pi_h$ for noisy-OR and SSSC and $\widetilde{s}=H\pi$ for BSC). We furthermore used an average bitflip probability of $p_{\mathrm{bf}}=\frac{1}{H}$.

\section{Data estimator}
\label{app:data_estimator}
\noindent{\it Models with binary latents and continuous observables.} Consider the posterior predictive distribution
\begin{align}
\label{eq:binlat_posterior_predictive_distribution}
p(\yVecEst\mid \yVecObs, \Theta) 
= \sum_{\sVec} p(\yVecEst, \sVec \mid \yVecObs, \Theta) 
= \sum_{\sVec} p(\yVecEst \mid \sVec, \Theta) p(\sVec \mid \yVecObs, \Theta)
\end{align}
The second step in \eqref{eq:binlat_posterior_predictive_distribution} exploits the fact that $\yVecEst$ and $\yVecObs$ are conditionally independent given the latents.
In order to infer the value of $\yDEst$ we will take expectations \wrt $p(\yVecEst\mid\yVecObs)$:
\begin{align}
\label{eq:posterior_predictive_distribution_expectation}
\langle \yDEst \rangle_{p(\yVecEst \mid \yVecObs, \Theta)} = \int\limits_{\yVecEst} \yDEst p(\yVecEst \mid \yVecObs, \Theta)\,\mathrm{d}\yVecEst = \int\limits_{\yDEst} \yDEst p(\yDEst \mid \yVecObs, \Theta)\,\mathrm{d}\yDEst.
\end{align}
The second step in \eqref{eq:posterior_predictive_distribution_expectation} follows from marginalizing over $\yVecEst\backslash\yDEst$. Equations \ref{eq:binlat_posterior_predictive_distribution} and \ref{eq:posterior_predictive_distribution_expectation} can be combined:
\begin{align}
\label{eq:binlat_general_data_estimator_start}
\langle \yDEst \rangle_{p(\yVecEst \mid \yVecObs, \Theta)} &= \int\limits_{\yDEst} \yDEst \sum_{\sVec} p(\yDEst \mid \sVec, \Theta) p(\sVec \mid \yVecObs, \Theta)\,\mathrm{d}\yDEst\\
&= \sum_{\sVec} \int\limits_{\yDEst} \yDEst p(\yDEst \mid \sVec, \Theta)\,\mathrm{d}\yDEst\,p(\sVec \mid \yVecObs, \Theta)\\
&= \sum_{\sVec} \langle \yDEst \rangle_{p(\yDEst \mid \sVec, \Theta)}\,p(\sVec \mid \yVecObs, \Theta)\\
\label{eq:binlat_general_data_estimator_final_copy}
&= \left \langle \langle \yDEst \rangle_{p(\yDEst \mid \sVec, \Theta)} \right \rangle_{p(\sVec \mid \yVecObs, \Theta)}.
\end{align}
The inner expectation in \eqref{eq:binlat_general_data_estimator_final_copy} is for the example of the BSC model \eqref{eq:bsc_gen_model} given by $\langle y_d \rangle_{p(y_d \mid \sVec, \Theta)} = \WdVec \sVec$ with $\WdVec$ denoting the $d$-th row of the matrix $W$. With this \eqref{eq:binlat_general_data_estimator_final_copy} takes for the BSC model the following form
\begin{equation}
\label{eq:bsc_data_estimator}
\langle \yDEst \rangle_{p(\yVecEst \mid \yVecObs, \Theta)} = \WdVec \langle\sVec\,\rangle_{p(\sVec\mid\yVecObs,\Theta)}.
\end{equation}
The estimator \eqref{eq:bsc_data_estimator} can be efficiently computed by (i) approximating the full posterior distribution of the latents by using truncated posteriors \eqref{eq:tvem_truncated_posterior} and by (ii) approximating the expectation \wrt the full posterior by applying truncated expectations \eqref{eq:tvem_truncated_expectations}. 
\\
\\
\noindent{\it Models with binary-continuous latents and continuous observables.} The upper derivation can be extended to be applicable to models with binary-continuous latents. Following the same line of reasoning, we again start with the posterior predictive distribution $p(\yVecEst\mid\yVecObs)$:
\begin{align}
p(\yVecEst\mid \yVecObs, \Theta) 
&= \sum_{\sVec} \int\limits_{\zVec} p(\yVecEst, \sVec, \zVec \mid \yVecObs, \Theta)\,\dz \\
\label{eq:bincontlat_posterior_predictive_distribution}
&=\sum_{\sVec} \int\limits_{\zVec} p(\yVecEst \mid \sVec, \zVec, \Theta) p(\sVec \mid \yVecObs, \Theta) p(\zVec \mid \sVec, \yVecObs, \Theta)\,\dz.
\end{align}
We then follow the steps from equations \ref{eq:binlat_general_data_estimator_start} to \ref{eq:binlat_general_data_estimator_final_copy}, i.e. we take expectations \wrt the posterior predictive distribution \eqref{eq:bincontlat_posterior_predictive_distribution}:
\begin{align}
\langle \yDEst \rangle_{p(\yVecEst \mid \yVecObs, \Theta)} &= \int\limits_{\yDEst} \yDEst \sum_{\sVec} \int\limits_{\zVec} p(\yDEst \mid \sVec, \zVec, \Theta) p(\sVec, \zVec \mid \yVecObs, \Theta)\,\dz\,\mathrm{d}\yDEst\\
&= \sum_{\sVec} \int\limits_{\zVec} \int\limits_{\yDEst} \yDEst p(\yDEst \mid \sVec, \zVec, \Theta)\,\mathrm{d}\yDEst\,p(\sVec, \zVec \mid \yVecObs, \Theta)\,\dz\\
&= \sum_{\sVec} \int\limits_{\zVec} \langle \yDEst \rangle_{p(\yDEst \mid \sVec, \zVec, \Theta)}\,p(\sVec, \zVec \mid \yVecObs, \Theta)\,\dz\\
\label{eq:bincontlat_general_data_estimator_final}
&= \left \langle \langle \yDEst \rangle_{p(\yDEst \mid \sVec, \zVec, \Theta)} \right \rangle_{p(\sVec, \zVec \mid \yVecObs, \Theta)}.
\end{align}
For the example of the SSSC model \eqref{eq:sssc_gen_model_prior}-\eqref{eq:sssc_gen_model_conditional}, the inner expectation in \eqref{eq:bincontlat_general_data_estimator_final} is given by $\langle y_d \rangle_{p(y_d \mid \sVec, \zVec, \Theta)} = \WdVec \szVecBr$ s.t. equation \ref{eq:bincontlat_general_data_estimator_final} takes for the SSSC model the following form:
\begin{equation}
\label{eq:sssc_data_estimator}
\langle \yDEst \rangle_{p(\yVecEst \mid \yVecObs, \Theta)} = \WdVec \langle\szVec\,\rangle_{p(\sVec,\zVec\mid\yVecObs,\Theta)}.
\end{equation}
The estimator \eqref{eq:sssc_data_estimator} can be efficiently computed based on \eqref{eq:sssc_exp_sz} by approximating the expectation \wrt the binary posterior by using truncated expectations \eqref{eq:tvem_truncated_expectations}.

\section{M-step update equations}
\label{app:mstep_equations}
{\it Noisy-OR.} For completeness, we report the Noisy-OR update rules here:
\begin{equation} 
  \pi_h  = \frac{1}{N} \sum_{n=1}^{N} \left< s_h \right>_{q^{(n)}}, ~~~~~~~~W _{dh} = 1 + \frac{\sum_{n=1}^{N} (y_d^{(n)} - 1) \left<D_{dh}(\sVec)\right>_{q^{(n)}}}{\sum_{n=1}^{N} \left<C_{dh}(\sVec)\right>_{q^{(n)}}}
\end{equation}
where
\begin{equation}
  \label{eq:nor_mstep_Wdefs}
  D_{dh}(\sVec) \defeq \frac{\widetilde{W}_{dh}(\sVec)s_h}{N_d(\sVec)(1-N_d(\sVec))}, \,\,\,C_{dh}(\sVec) \defeq \widetilde{W}_{dh}(\sVec)D_{dh}(\sVec), \,\,\,\widetilde{W}_{dh}(\sVec) \defeq \prod_{h' \neq h}(1 - W_{dh'}s_{h'})
\end{equation}
The update equations for the weights $W_{dh}$ do not allow for a closed-form solution. We instead employ a fixed-point equation whose fixed point is the exact solution of the maximization step. We exploit the fact that in practice one single evaluation of \eqref{eq:nor_mstep_Wdefs} is enough to (noisily, not optimally) move towards convergence
to efficiently improve on the parameters $W_{dh}$.\\

\noindent{\it BSC.} As for noisy-OR, we report the explicit forms of the M-step update rules for the BSC model here for completeness \citep[compare, e.g.,][]{HennigesEtAl2010}:
\begin{equation}
\begin{gathered}
\pi = \frac{1}{NH} \sum_{n=1}^{N}\sum_{h=1}^{H} \left<s_h\right>_{q^{(n)}}, ~~~~~~~~ \sig^2 = \frac{1}{ND} \sum_{n=1}^{N} \left<\vert\vert\,\yVecN-W\sVec\,\vert\vert^2\right>_{q^{(n)}}\\
W = \left(\sum_{n=1}^{N}\yVecN\left<\sVec\,\right>_{q^{(n)}}^T\right)\left(\sum_{n'=1}^{N}\left<\sVec\sVec^{\,T}\right>_{q^{(n')}}\right)^{-1}
\end{gathered}
\end{equation}\\

\noindent{\it SSSC.} We report the final expressions of the SSSC M-step update equations below. The derivations can be found in \citet{SheikhEtAl2014}.
\begin{equation}
\label{eq:sssc_mstep}
\begin{gathered}
  W = \frac{ \sum_{n=1}^{N} \yVecN \bigl<\szVec\,\bigr>_{\qn}\transp }{\sum_{n=1}^N \bigl<\szVecBr \szVecBr\transp\bigr>_{\qn} }, ~~~~~~ \piVec = \frac{1}{N} \sum_{n=1}^{N} \bigl<\sVec\,\bigr>_{\qn}, ~~~~~~ \muVec = \frac{ \sum_{n=1}^{N} \bigl<\szVec\,\bigr>_{\qn} }{\sum_{n=1}^{N} \bigl<\sVec\,\bigr>_{\qn} }, \\
  \Psi = \sum_{n=1}^{N} \left[ \bigl<\szVecBr \szVecBr\transp \bigr>_{\qn}  - \bigl<\sVec \sVec\transp\bigr>_{\qn} \odot \muVec \muVec\transp \right] \odot \left(\sum_{n=1}^{N} \left[ \bigl<\sVec \sVec\transp \bigr>_{\qn} \right] \right)^{-1},\\
  \sigma^2 = \frac{1}{ND} \Tr \left(\sum_{n=1}^{N} \left[\yVecN (\yVecN)\transp - W\left[\bigl<\szVec\,\bigr>_{\qn} \bigl<\szVec\,\bigr>_{\qn}\transp \right] W\transp \right]\right)
\end{gathered}
\end{equation}
As defined in \eqref{eq:sssc_exp_values} the expectations in \eqref{eq:sssc_mstep} are taken \wrt the full binary-continuous latent space. Importantly for applying EEM, all these expectation values can be reformulated as expectations \wrt the posterior over the binary latent space:
\begin{align} 
\label{eq:sssc_exp_s}
\bigl<\sVec\,\bigr>_{\qn} &= \sum_{\sVec} \qn (\sVec;\Theta)\sVec\\
\label{eq:sssc_exp_ssT}
\bigl<\sVec\sVec\transp\bigr>_{\qn} &= \sum_{\sVec} \qn (\sVec;\Theta)\sVec\sVec\transp\\
\label{eq:sssc_exp_sz}
\bigl<\szVec\,\bigr>_{\qn} &= \sum_{\sVec} \qn (\sVec;\Theta)\vec{\kappa}_{\sVec}^{(n)}\\
\label{eq:sssc_exp_szszT}
\bigl<\szVecBr\szVecBr\transp\bigr>_{\qn} &= \sum_{\sVec} \qn (\sVec;\Theta) (\Lambda_{\sVec}+\vec{\kappa}_{\sVec}^{(n)}(\vec{\kappa}_{\sVec}^{(n)})\transp)
\end{align}
where $\qn(\sVec;\Theta)$ is (for the exact EM solution) the binary posterior given by 
\begin{gather} \label{eq:sssc_posterior_s}
  \posterior = \frac{\BBernou(\sVec;\piVec) \,\, \NGauss(\yVec;\,\tWs\, \muVec,\Cs)}{\sum_{\sVec'} \BBernou(\sVec\,';\piVec) \,\, \NGauss(\yVec;\,\tilde{W}_{\sVec'} \, \muVec,C_{\sVec'}) }.
\end{gather}
In the above equations $\Lambda_{\sVec}=( \sigma^{-2}\tilde{W}_{\vec{s}}\transp \tilde{W}_{\vec{s}} + \Psi_{\vec{s}}^{-1} )^{-1}$ and $\vec{\kappa}^{(n)}_{\sVec}=(\vec{s}\odot\vec{\mu}) + \sigma^{-2}\Lambda_{\vec{s}}\tilde{W}_{\vec{s}}\transp (\vec{y}^{\,(n)}-\tilde{W}_{\vec{s}}\vec{\mu})$. $\sum_{\sVec}$ denotes a summation over all binary vectors $\sVec$.
Based on the reformulations, the expectations in \eqref{eq:sssc_exp_s}-\eqref{eq:sssc_exp_szszT} can be approximated using truncated expectations \eqref{eq:tvem_truncated_expectations}.

\section{Technical details on numerical experiments}
\label{app:numerical_experiments_tech}
\noindent{\it Soft- and Hardware.} We implemented our algorithms in Python and used the \lstinline{mpi4py} package \citep{DalcinEtAl2005} for parallel execution on multiple processors. Model parameter updates were efficiently computed by distributing data points to multiple processors and performing calculations for batches of data points in parallel (compare Appendix\,\ref{app:mstep_equations}) .
Small scale experiments such as the bars tests described in Section\,\ref{subsec:verification} were performed on standalone workstations with four-core CPUs; runtimes were in the range of minutes. For most of the large-scale experiments described in Sections\,\ref{subsec:scalability} and \ref{sec:performance}, we used the HPC cluster CARL of the University of Oldenburg to execute our algorithms. The cluster is equipped with several hundred compute nodes with in total several thousands of CPU cores (Intel Xeon E5-2650 v4 12C, E5-2667 v4 8C and E7-8891 v4 10c). 
For each of the simulations, potentially different compute ressources were employed (different numbers of CPU cores, different CPU types, different numbers of compute nodes, different numbers of employed cores per node) and hence runtimes were found to vary greatly between the simulations. For example for the denoising experiments with EBSC and ES3C on the ``House'' image, the algorithms were executed using between one hundred and 640 CPU cores, and runtimes ranged approximatly between six and one hundred hours for one run. Images larger than ``House'' required still longer runtimes
or more CPU cores. 
For parts of the inpainting experiments (Sec.\,\ref{subsec:inpainting}), computing resources of the ``Norddeutscher Verbund zur F{\"o}rderung des Hoch- und H{\"o}chstleistungsrechnens'' (HLRN) were available. These resources allowed distributed execution of the algorithms on several thousand CPU cores (as opposed to several hundred cores as on the CARL cluster) and thus enabled significantly shorter runtimes of the algorithms as compared to the runtimes on the CARL cluster. 
\\ \\
%
\noindent{\it Hyperparameters.} Table\,\ref{tab:hyperparams} lists the hyperparameters employed in the numerical experiments on verification (Section\,\ref{subsec:verification}), scalability (Section\,\ref{subsec:scalability}), denoising (Section\,\ref{subsec:denoising}) and inpainting (Section\,\ref{subsec:inpainting}). EEM hyperparameters ($S$, $N_p$, $N_m$, $N_g$) were chosen s.t. they provided a reasonable tradeoff between the accuracy of the approximate inference scheme and the runtime of the algorithm.
\begin{table}[h!]
\centering
\resizebox{\linewidth}{!}{
\begin{tabular}{ c }
\begin{tabular}[c c c c c c c c c c]{c c c c c c c c c c}
\multicolumn{10}{l}{{\Large\bf A} Verification and scalability (Sections\,\ref{subsec:verification} and \ref{subsec:scalability})}\\
\toprule
Experiment&$N$&$D$&$H$&EA&$S$&$N_p$&$N_m$&$N_g$&Iterations\\
\midrule
Bars Test NOR/BSC/SSSC&5,000&$5\times5$&10&randparents-randflips&20&5&4&2&300\\
Bars Test NOR/BSC/SSSC&5,000&$5\times5$&10&fitparents-cross-randflips&20&5&-&2&300\\
Bars Test NOR/BSC/SSSC&5,000&$5\times5$&10&fitparents-sparseflips&20&5&4&2&300\\
Bars Test NOR/BSC/SSSC&5,000&$5\times5$&10&fitparents-cross-sparseflips&20&5&-&2&300\\
Bars Test NOR/BSC/SSSC&5,000&$5\times5$&10&randparents-cross-sparseflips&20&5&-&2&300\\
Bars Test NOR/BSC/SSSC&5,000&$5\times5$&10&fitparents-randflips&20&5&4&2&300\\
van Hateren Images NOR&30,000&$10\times10$&100&fitparents-cross-sparseflips&120&8&-&2&200\\
van Hateren Images BSC&100,000&$16\times16$&300&fitparents-cross-sparseflips&200&10&-&4&4,000\\
van Hateren Images SSSC&100,000&$12\times12$&512&fitparents-cross-sparseflips&60&6&-&2&2,000\\
\bottomrule
\end{tabular}\\ \\ \\
\begin{tabular}[c c c c c c c c c c c]{c c c c c c c c c c c}
\multicolumn{11}{l}{{\Large\bf B} Denoising (Relation between PSNR measure and variational lower bound; Section\,\ref{subsec:denoising_psnr_free_energy})}\\
\toprule
Image&Size&Noise Level $\sigma$&$D$&$H$&EA&$S$&$N_p$&$N_m$&$N_g$&Iterations\\
\midrule
House&$256\times256$&50&$8\times8$&256&fitparents-randflips&60&6&5&2&2,000\\
\bottomrule
\end{tabular}\\ \\ \\
\begin{tabular}[c c c c c c c c c c c]{c c c c c c c c c c c}
\multicolumn{11}{l}{{\Large\bf C} Denoising (Comparison of generative models and approximate inference under controlled conditions; Section\,\ref{subsec:denoising_controlled_conditions})}\\
\toprule
Image&Size&Noise Level $\sigma$&$D$&$H$&EA&$S$&$N_p$&$N_m$&$N_g$&Iterations\\
\midrule
House&$256\times256$&50&$8\times8$&64&fitparents-randflips&60&60&1&1&4,000\\
House&$256\times256$&50&$8\times8$&256&fitparents-randflips&60&60&1&1&4,000\\
House&$256\times256$&50&$12\times12$&512&fitparents-randflips&60&60&1&1&4,000\\
\bottomrule
\end{tabular}\\ \\ \\
\begin{tabular}[c c c c c c c c c c c c c c c c c]{c c c c c c c c c c c c c c c c c}
\multicolumn{17}{l}{{\Large\bf D} Denoising (General comparison of denoising approaches; Section\,\ref{subsec:denoising_general_benchmarking})}\\
\toprule
\multirow{2}{*}{Image}&\multirow{2}{*}{Size}&\multirow{2}{*}{Noise Level $\sigma$}&\multicolumn{6}{c}{Hyperparameters EBSC}&\multicolumn{6}{c}{Hyperparameters ES3C}&\multirow{2}{*}{EA}&\multirow{2}{*}{Iterations}\\
\cmidrule(lr){4-9}
\cmidrule(lr){10-15}
&&&$D$&$H$&$S$&$N_p$&$N_m$&$N_g$&$D$&$H$&$S$&$N_p$&$N_m$&$N_g$&\\
\midrule
House&$256\times256$&15/25/50&$8\times8$&256&200&10&9&4&$12\times12$&512&60&60&1&1&fitparents-randflips&4,000\\
Barbara&$512\times512$&25&$8\times8$&256&200&10&9&4&$11\times11$&512&60&60&1&1&fitparents-randflips&3,000\\
Lena&$512\times512$&25&$8\times8$&256&200&10&9&4&$11\times11$&512&60&60&1&1&fitparents-randflips&4,000\\
Peppers&$256\times256$&25&$8\times8$&256&200&10&9&4&$10\times10$&800&40&30&1&1&fitparents-randflips&6,000\\
\bottomrule
\end{tabular}\\ \\ \\
\begin{tabular}[c c c c c c c c c c c]{c c c c c c c c c c c}
\multicolumn{11}{l}{{\Large\bf E} Inpainting (Section\,\ref{subsec:inpainting})}\\
\toprule        
Image&Size&Missing Data Ratio&$D$&$H$&EA&$S$&$N_p$&$N_m$&$N_g$&Iterations\\
\midrule
Barbara&$512\times512$&50\%&$12\times12$&512&fitparents-randflips&30&20&1&1&4,000\\
Cameraman&$256\times256$&50\%&$12\times12$&512&fitparents-randflips&30&20&1&1&4,000\\
Lena&$512\times512$&50\%&$12\times12$&512&fitparents-randflips&30&20&1&1&4,000\\
House&$256\times256$&50\%&$12\times12$&512&fitparents-randflips&30&20&1&1&4,000\\
House&$256\times256$&80\%&$15\times15$&512&fitparents-randflips&30&20&1&1&500\\
Castle&$481\times321$&50\%&$7\times7$&900&fitparents-randflips&30&20&1&1&2,000\\
Castle&$481\times321$&80\%&$7\times7$&900&fitparents-randflips&60&60&1&1&200\\
New Orleans&$297\times438$&Text Mask&$14\times14$&900&fitparents-randflips&60&60&1&1&3,000\\
\bottomrule
\end{tabular}\\
\end{tabular}
}
\caption{Hyperparameters employed in the numerical experiments.}
\label{tab:hyperparams}
\end{table}

\section{Further numerical results}
\label{app:numerical_results}
\begin{figure*}[!hbt]
\includegraphics[]{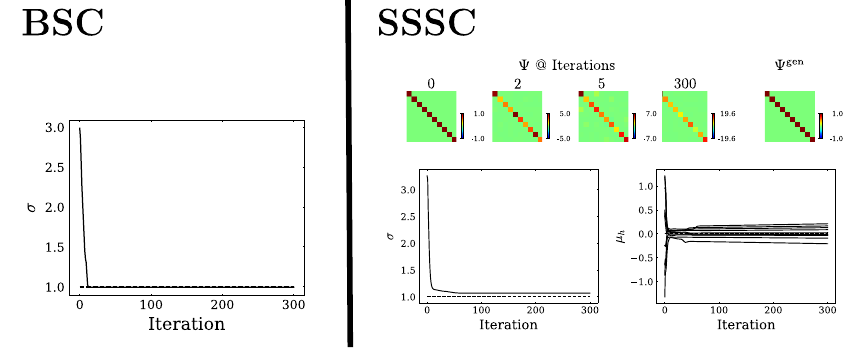}
\caption{BSC and SSSC model parameters learned from artificial data using EEM (for more details see the description of the experiment in Section\,\ref{subsec:verification}).}
\vspace{-.1in}
\label{fig:app_bscssscBars}
\end{figure*}

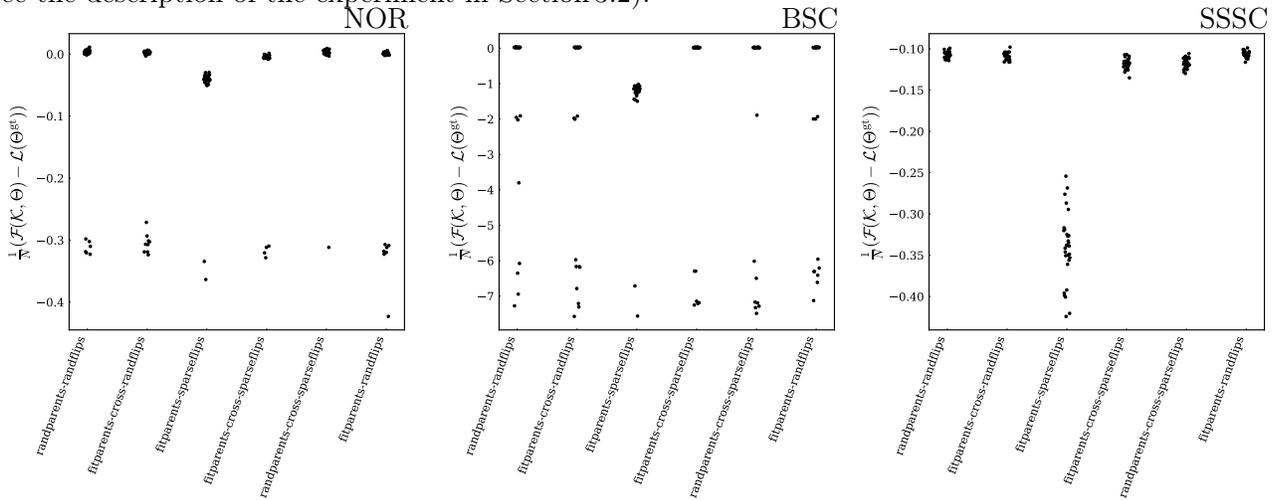
\begin{figure*}[!hbt]
\vspace{-.1in}
\centering
\newcommand\XP{0.01cm}
\newcommand\XF{0.32\textwidth}
\newcommand\YL{0.01cm}
\newcommand\XL{0.0cm}
\begin{tikzpicture}
\node(a) at (0,0) {\includegraphics[width=\XF]{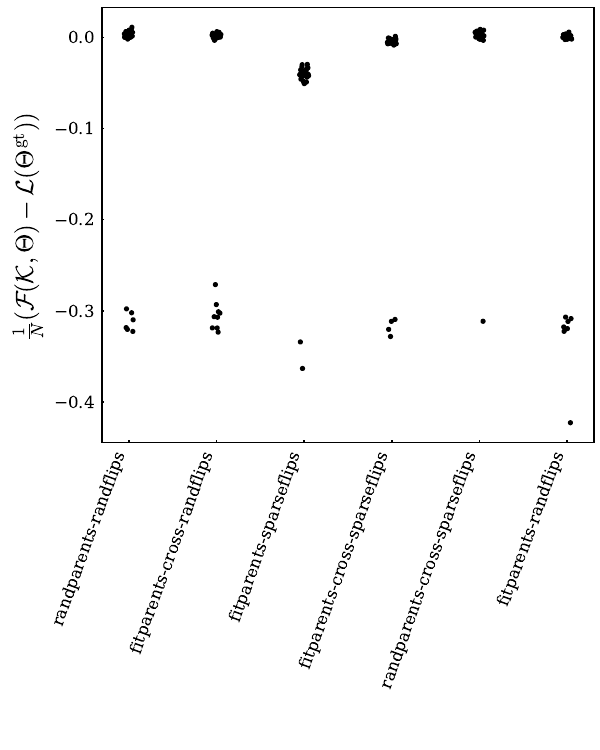}};
\node[right = \XP of a] (b) {\includegraphics[width=\XF]{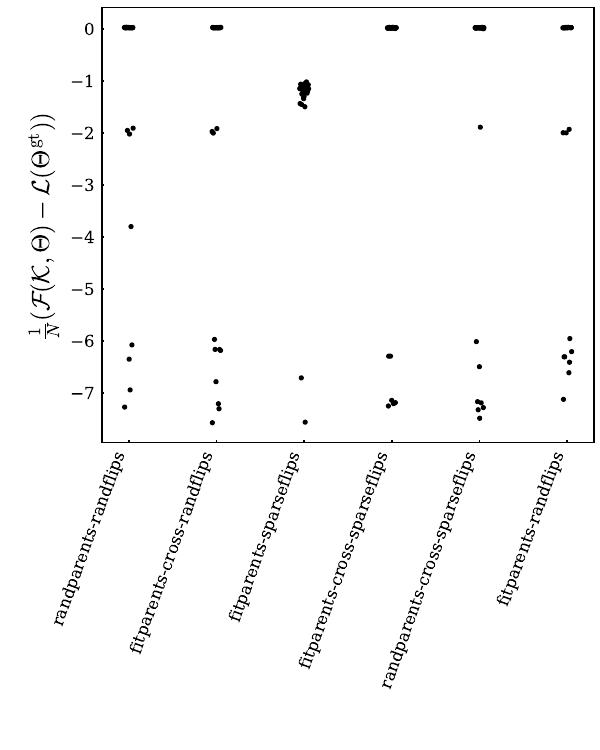}};
\node[right = \XP of b](c) {\includegraphics[width=\XF]{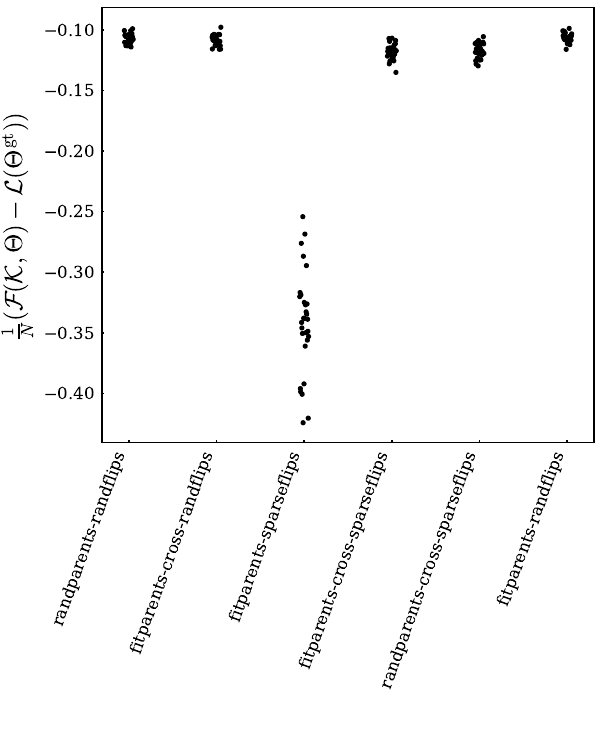}};
\node[above left = \YL and \XL of a.north east, anchor = east] (al) {NOR};
\node[above left = \YL and \XL of b.north east, anchor = east] (bl) {BSC};
\node[above left = \YL and \XL of c.north east, anchor = east] (cl) {SSSC};
\end{tikzpicture}
\caption{Final free energies obtained by different EA configurations over 30 runs of EEM for noisy-OR, BSC and SSSC on 5$\times$5 bars images (for more details see the description of the experiment in Section\,\ref{subsec:verification}).}
\vspace{-.1in}
\label{fig:app_norbscssscReliability}
\end{figure*}

\begin{figure*}
\centering
\includegraphics[]{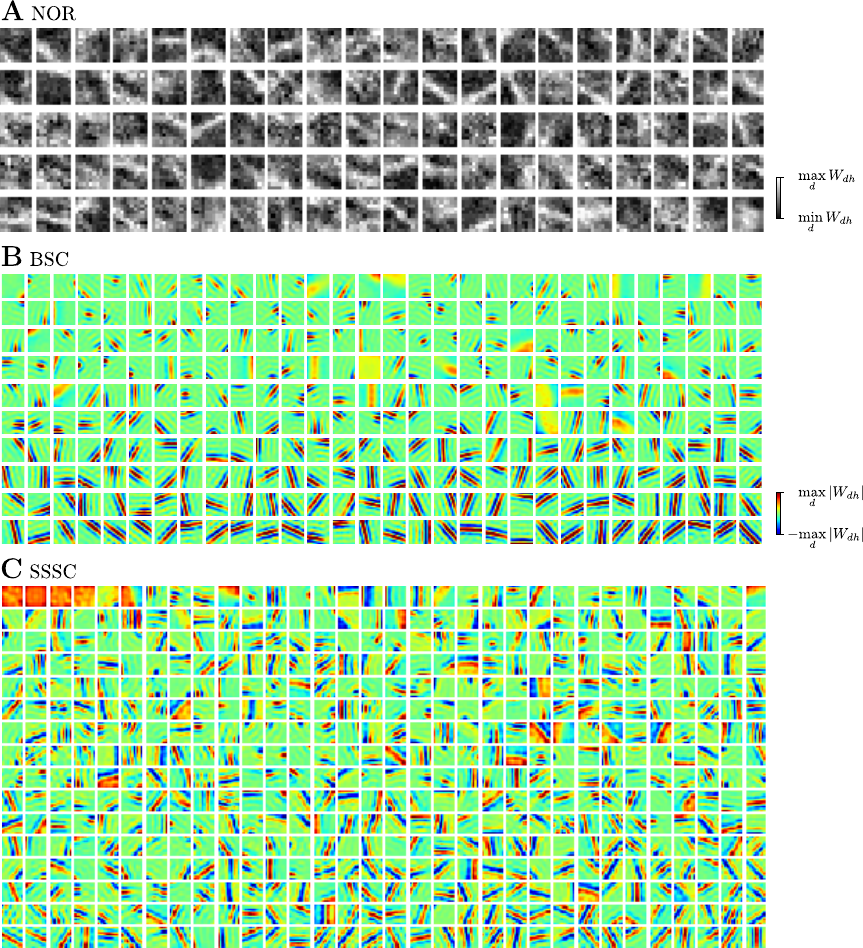}
\caption{Dictionaries learned from natural image patches using noisy-OR, BSC and SSSC models (see Section\,\ref{subsec:scalability}). For noisy-OR, we considered raw image patches and trained a model with $H=100$ components. For BSC and SSSC, image patches were preprocessed using a whitening procedure. For BSC and SSSC, $H=300$ and $H=512$ generative fields were learned, respectively. In {\bf B} and {\bf C}, the fields are ordered according to their activation, starting with the fields corresponding to the most active hidden units.}
\vspace{-.1in}
\label{fig:app_norbscssscHateren}
\end{figure*}

\begin{figure*}[!hbt]
\centering
\includegraphics{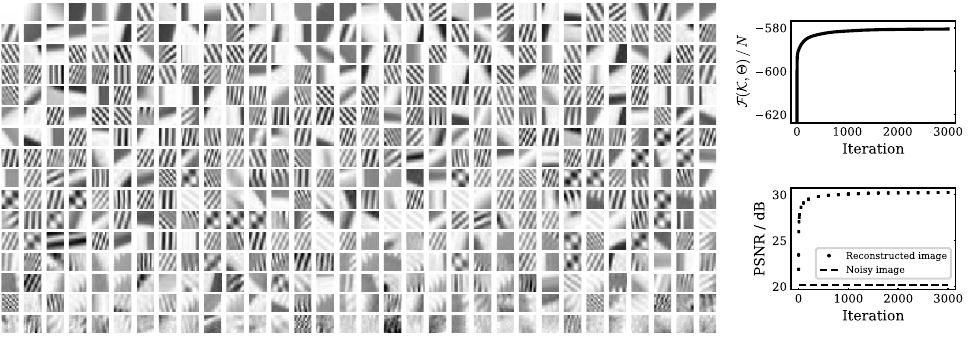};
\caption{SSSC generative fields learned from the noisy ``Barbara'' image (compare Section\,\ref{subsec:denoising}). The fields are ordered according to their activation, starting with the fields corresponding to the most active hidden units.}
\label{fig:app_ssscBarbara}
\end{figure*}

Figures\,\ref{fig:app_bscssscBars} to \ref{fig:app_ssscBarbara} provide further results related to the numerical experiments described in Sections\,\ref{sec:applications}\,-\,\ref{sec:performance}.

\begin{table*}[!hbt]
\centering
\resizebox{\linewidth}{!}{
\begin{tabular}{ c }
\begin{tabular}[c c c c c c c c c c c c c]{c c c c c c c c c c c c c}
  \multicolumn{13}{l}{{\Large\bf A} Denoising EBSC and ES3C}\\
  \toprule
  &\multicolumn{2}{c}{House $\sigma=15$}&\multicolumn{2}{c}{House $\sigma=25$}&\multicolumn{2}{c}{House $\sigma=50$}&\multicolumn{2}{c}{Barbara $\sigma=25$}&\multicolumn{2}{c}{Lena $\sigma=25$}&\multicolumn{2}{c}{Peppers $\sigma=25$}\\
  \cmidrule(lr){2-3}
  \cmidrule(lr){4-5}
  \cmidrule(lr){6-7}
  \cmidrule(lr){8-9}
  \cmidrule(lr){10-11}
  \cmidrule(lr){12-13}
  &EBSC&ES3C&EBSC&ES3C&EBSC&ES3C&EBSC&ES3C&EBSC&ES3C&EBSC&ES3C\\
  \midrule
  Run 1&33.70&34.86&32.39&33.16&29.03&29.88&28.33&30.17&30.91&31.90&29.01&30.35\\
  Run 2&33.65&34.93&32.47&33.20&28.91&29.85&28.39&30.14&30.90&31.94&28.87&30.22\\
  Run 3&33.63&34.93&32.34&33.09&29.01&29.74&28.42&30.24&30.84&32.01&28.90&30.26\\
  \midrule
  Average&33.66$\pm$0.03&34.90$\pm$0.03&32.40$\pm$0.05&33.15$\pm$0.05&28.98$\pm$0.05&29.83$\pm$0.06&28.38$\pm$0.04&30.19$\pm$0.04&30.88$\pm$0.03&31.95$\pm$0.04&28.93$\pm$0.06&30.27$\pm$0.06\\
  \bottomrule
\end{tabular}\\ \\ \\
\begin{tabular}[c c c c c c c c]{c c c c c c c c}
  \multicolumn{8}{l}{{\Large\bf B} Inpainting ES3C}\\
  \toprule
  &Barbara&Cameraman&Lena&\multicolumn{2}{c}{House}&\multicolumn{2}{c}{Castle}\\
  \cmidrule(lr){2-2}
  \cmidrule(lr){3-3}
  \cmidrule(lr){4-4}
  \cmidrule(lr){5-6}
  \cmidrule(lr){7-8}
  &50\% missing&50\% missing&50\% missing&50\% missing&80\% missing&50\% missing&80\% missing\\
  \midrule
  Run 1&35.39&30.91&37.57&39.55&32.99&38.14&29.65\\
  Run 2&35.51&30.46&37.47&39.70&33.21&38.39&29.68\\
  Run 3&35.42&30.70&37.58&39.54&32.99&38.14&29.64\\
  \midrule
  Average&35.44$\pm$0.05&30.69$\pm$0.18&37.54$\pm$0.05&39.59$\pm$0.07&33.06$\pm$0.10&38.23$\pm$0.12&29.66$\pm$0.02\\
  \bottomrule
\end{tabular}
\end{tabular}
}
\caption{PSNR values (in dB) measured for EBSC and ES3C in the denoising and inpainting experiments described in Sections\,\ref{subsec:denoising_general_benchmarking} and \ref{subsec:inpainting}. Panel A is related to the results shown in Figures\,\ref{fig:ssscDenoising_house_all_in_one}\,D and \ref{fig:ssscDenoising_barblenpep}; panel B is related to the data of Figure\,\ref{fig:ssscInpainting}\,A-E.}
\label{tab:app_den_inp_all_runs}
\end{table*}

\section{Evaluation Criteria}
\label{app:discussion_on_evaluation}
A standard metric for the evaluation of image restoration (IR) methods is the peak-signal-to-noise ratio (PSNR; compare Section\,\ref{sec:performance}). In the context of IR benchmarks (e.g., denoising, inpainting, image super resolution), PSNR values are informative about the root-mean-square error between the restored image calculated by a specific IR algorithm and the respective clean target image. In addition to the PSNR measure, there are other criteria that can be used to compare IR approaches. Some criteria will be discussed in the following.
\\
\\
\noindent{\it Clean data.}
Supervised learning-based IR methods such as the denoising and inpainting methods of category DE6 and respectively IN6 in Table\,\ref{tab:requirements} require external datasets with clean images for training. In contrast, sparse coding and dictionary learning approaches such as the methods from category DE1 and IN1 (including the generative model algorithms EBSC and ES3C) are trained exclusively on the corrupted data they aim to restore. The ability of learning solely from corrupted data is very valuable for scenarios in which clean data is not available or difficult to generate.
Besides, the internal statistics of a test image were observed to be often more predictive than statistics learned from external datasets \citep[compare][]{ShocherEtAl2018}.
For deep neural networks, which often do require external clean training data, recent work seeks to provide methods to also allow them to be trained on noisy data alone \citep[e.g.][]{LehtinenEtAl2018,UlyanovEtAl2018,KrullEtAl2019}. Further related work proposes methods for ``zero-shot'' super-resolution which can be trained exclusively on the corrupted test data \citep{ShocherEtAl2018}.
\\
\\
\noindent{\it A priori information and robustness.}
For denoising, it is frequently assumed that a priori knowledge of the ground-truth noise level is available. For instace KSVD, BM3D, WNNM, cKSVD or LSSC treat the noise level as input parameter of the algorithm. Other approaches that are trained using pairs of noisy/clean examples are typically optimized either for a single noise level (e.g. MLP, IRCNN, DnCNN-S, TNRD) or for several noise levels (e.g., MemNet, DnCNN-B, FFDNet, DPDNN, BDGAN). The denoising performance of noise-level-optimized approaches may deteriorate significantly if the algorithm is not provided with the appropriate (ground-truth) noise level of the test image \citep[compare, e.g.,][]{BurgerEtAl2012,ChaudhuryAndRoy2017,ZhangEtAl2018}.

Similarly, for inpainting, it can be observed that certain methods exploit significant amounts of a priori information, for example considering the text removal benchmark (Figure\,\ref{fig:ssscInpainting}\,F): For NLRMRF and IRCNN, the training data is generated by overlaying specific text patterns on clean images. For IRCNN, several distinct font styles and font sizes are used \citep[][]{ChaudhuryAndRoy2017}; for NLRMRF, training examples are generated using the identical text pattern of the test image \citep[][]{SunAndTappen2011}. 
In contrast, the generative model algorithms EBSC and ES3C require neither task-specific external training datasets (such as IRCNN or NLRMRF) nor do they require a-priori knowledge of the noise level (in contrast to e.g. BM3D, MLP, IRCNN, FFDNET, DnCNN; compare Table\,\ref{tab:requirements}). EBSC and ES3C learn the noise level parameter (which is part of the generative model) from the data in an unsupervised way.
\\
\\
\noindent{\it Context information.} 
Another criterion for the evaluation of IR methods is the amount of context information that a particular algorithm exploits. The amount of context information is primarily determined by the patch size of the image segmentation. Increasing the effective patch size was reported to be beneficial for the performance of IR algorithms \citep[compare][]{BurgerEtAl2012,ZhangEtAl2017}. 
Zhang et al. reported that the effective patch size used by denoising methods can be found to vary greatly between different approaches (e.g. $36\times36$ used by EPLL, $50\times50$ by DnCNN-B, $70\times70$ by FFDNet, $361\times361$ by WNNM; numbers taken from \citealp{ZhangEtAl2017,ZhangEtAl2018}). 
The denoising experiments with EBSC and ES3C were conducted using patches that did not exceed an effective size of $23\times23$ (compare Table~\ref{tab:hyperparams} in Appendix\,\ref{app:numerical_experiments_tech}) which is considerably smaller than the numbers reported by Zhang et al.. 
\\
\\
\noindent{\it Stochasticity in the acquisition of test data.} For the denoising benchmarks considered in Section\,\ref{subsec:denoising}, there is stochastic variation in the test data due to the fact that for a given AWG noise level $\sigma$ and for a given image, different realizations of the noise result in different noisy images. PSNR values can consequently vary even if the applied algorithm is fully deterministic. 
As all images investigated here are at least of size $256\times{}256$, variations due to different noise realizations are commonly taken as negligible for comparison \citep[see, e.g.,][]{MairalEtAl2009}. 
The denoising results of EBSC and ES3C reported in Figures\,\ref{fig:ssscDenoising_house_all_in_one} and \ref{fig:ssscDenoising_barblenpep} were obtained by executing the algorithm three times on each image using different noise realizations in each run. Observed PSNR standard deviations were smaller or equal 0.06\,dB (compare Table\,\ref{tab:app_den_inp_all_runs}). For the inpainting experiments with randomly missing values (Figure\,\ref{fig:ssscInpainting}\,A-E), we also performed three runs for each image using a different realization of missing values in each run. In these experiments, we observed slightly higher PSNR variations (standard deviations ranged from 0.02 to 0.18\,dB). 
The different realizations of the noise (or of the missing values) employed for each execution of the algorithm might be relevant to explain the PSNR variations we observed. Stochasticity in the EBSC and ES3C algorithms themselves however cannot be excluded as an additional factor. 
\\
\\
\noindent{\it Stochasticity in the algorithm.} The output of stochastic algorithms can differ from run to run even if the input remains constant. For learning algorithms, different PSNR values usually correspond to different optima of the learning objective which are obtained due to different initializations and/or due to stochasticity during learning. For stochastic algorithms, it is therefore the question which PSNR value shall be used for comparison. Only two of the algorithms we used for comparison in Figures\,\ref{fig:ssscDenoising_house_all_in_one}\,-\,\ref{fig:ssscInpainting} report averages over different runs (KSVD and LSSC). All other algorithms do report a single PSNR value per denoising/inpainting setting without values for standard deviations. A single PSNR value is (for one realization of the corrupted image) naturally obtained for deterministic algorithms (e.g. BM3D, WNNM, NL).
For the stochastic algorithms, a single PSNR may mean that either (A)~just one run of the learning algorithm was performed, that (B)~the best PSNR of all runs was reported, or that (C)~one run of the learning algorithm was selected via another criterion. For DNN algorithms, for instance, the DNN with the lowest training or validation error could be selected for denoising or inpainting. Or for sparse coding algorithms, the model parameters with the lowest (approximate) likelihood could be selected for denoising or inpainting. 
As the contributions in Figure\,\ref{fig:ssscDenoising_house_all_in_one}\,-\,\ref{fig:ssscInpainting} which state just one PSNR value do not give details on how it was obtained from potentially several runs, it may be assumed that the best performing runs were reported (or sometimes the only run, e.g., in cases of a DNN requiring several days or weeks for training, compare \citealt[][]{JainAndSeung2009,BurgerEtAl2012,ChaudhuryAndRoy2017,ZhangEtAl2018}). Stating the best run is instructive as it shows how well an algorithm can perform under best conditions. For comparability, it should be detailed how the single PSNR value was selected, however (or if average and best values are essentially identical).

For practical applications, it is desirable to be able to select the best of several runs based on a criterion that can be evaluated without ground-truth knowledge. 
Our experimental data shows that for EBSC and ES3C the best denoising and inpainting performance in terms of PSNR cannot reliably be determined based on the learning objective, namely the variational bound (which is computable without ground-truth knowledge): The PSNR values of the runs with the highest bound may be smaller than the highest PSNR values observed in all runs. 
In a scenario with fixed noise realization, we observed the variational bound to be instructive about the PSNR value though (Figure\,\ref{fig:ssscDenoising_house_psnr_fenergy}).

\section{Comparison to GANs and VAEs}
\label{app:comparison_to_gans_vaes}
While Figures\,\ref{fig:ssscDenoising_house_all_in_one} to \ref{fig:ssscInpainting} show a comparison of EBSC and ES3C with a range of other denoising and inpainting approaches, no competitive performance for GANs and VAEs, except for BDGAN, is shown. Recent VAE versions instead frequently report approximate log-likelihoods on (binarized or standard) MNIST or human face datasets such as CelebA \citep[e.g.,][]{ImEtAl2017,CreswellAndBharath2018}. \citeauthor{CreswellAndBharath2018} also report reconstruction performance on AWG noise removal tasks, however in \citet{CreswellAndBharath2018} (i) the considered noise levels are significantly smaller than the ones in Figures\,\ref{fig:ssscDenoising_house_all_in_one}\,-\,\ref{fig:ssscDenoising_barblenpep} and (ii) datasets such as Omniglot or CelebA are employed rather than the standard test images of Figures\,\ref{fig:ssscDenoising_house_all_in_one}\,-\,\ref{fig:ssscDenoising_barblenpep}.
Related contributions on GANs frequently consider the task of semantic inpainting, i.e. the reconstruction of large areas in images in a semantically plausible way \citep[see, e.g.,][for an overview]{CaoEtAl2018}. In these contributions, typically datasets such as CelebA, Street View House Numbers images or ImageNet are employed and frequently the reconstruction of a single or a few (e.g. squared or rectangular) holes is considered as a benchmark \citep[compare, e.g.,][]{PathakEtAl2016,YangEtAl2017,YehAtAl2017,IizukaEtAl2017,LiEtAl2017}. \citeauthor[][]{YehAtAl2017} also consider the task of restoring randomly missing values, however neither \citet[][]{YehAtAl2017} nor the aforementioned publications report performance on the test images of Figure\,\ref{fig:ssscInpainting}.
The benchmarks of denoising and inpainting we use (Figures\,\ref{fig:ssscDenoising_house_all_in_one} to \ref{fig:ssscInpainting}) are (A)~very natural for EEM because of available benchmark results for mean field and
 sampling approaches, and (B)~the benchmarks allow for comparison with state-of-the-art feed-forward DNNs (Figures\,\ref{fig:ssscDenoising_house_all_in_one} and \ref{fig:ssscDenoising_barblenpep}).

\end{appendices}





\end{document}